\def\cvprPaperID{5149}
\def\confName{ICCV}
\def\confYear{2023}

\def\paperTitle{VidStyleODE: Disentangled Video Editing via StyleGAN and NeuralODEs}

\def\authorBlock{
    Moayed Haji Ali\thanks{Equal contribution} \qquad
    Andrew Bond\footnotemark[1] \qquad
     \\
    Koç University \\
    {\tt\small \{mali18, abond19\}@ku.edu.tr}
}

\newif\ifreview 
\newif\ifarxiv \newcommand{\arxiv}{\arxivtrue}
\newif\ifcamera 
\newif\ifrebuttal 

\arxiv

\pdfoutput=1
\documentclass[10pt,twocolumn,letterpaper]{article}
\ifreview \usepackage[review]{cvpr} \fi
\ifarxiv \usepackage[pagenumbers]{cvpr} \fi
\ifrebuttal \usepackage[rebuttal]{cvpr} \fi
\ifcamera \usepackage{cvpr} \fi

\usepackage{graphicx}
\usepackage{amsmath}
\usepackage{amssymb}
\usepackage{booktabs}
\usepackage{comment}

\usepackage{times}
\usepackage{tabularx}
\usepackage{microtype}
\usepackage{epsfig}
\usepackage[table,xcdraw]{xcolor}
\usepackage{caption}
\usepackage{float}
\usepackage{placeins}
\usepackage{color, colortbl}
\usepackage{stfloats}
\usepackage{enumitem}
\usepackage{tabularx}
\usepackage{xstring}
\usepackage{multirow}
\usepackage{xspace}
\usepackage{url}
\usepackage{subcaption}
\usepackage{xcolor}
\usepackage{slashbox}
\usepackage[hang,flushmargin]{footmisc}

\ifcamera \usepackage[accsupp]{axessibility} \fi

\definecolor{darkgreen}{rgb}{0.0, 0.26, 0.15}

\newcommand{\blue}[1]{{\textcolor{blue}{[#1]}}}
\newcommand{\red}[1]{{\textcolor{red}{[#1]}}}
\newcommand{\orange}[1]{{\textcolor{orange}{[#1]}}}

\newcommand{\z}{\mathbf{z}}

\newcommand{\Vid}{\mathcal{V}}
\newcommand{\loss}{\mathcal{L}}
\newcommand{\R}{\mathbb{R}}
\newcommand{\Wplus}{\mathcal{W}_+}

\DeclareMathOperator*{\Expect}{\mathbb{E}}

\newcommand{\fmot}{f_D} 
\newcommand{\fmerge}{f_{G}}
\newcommand{\Dsrc}{\D_{\mathrm{SRC}}}
\newcommand{\Dtgt}{\D_{\mathrm{TGT}}}
\newcommand{\CLIP}{\mathrm{CLIP}}
\newcommand{\dz}{\Delta_\mathbf{z}}
\newcommand{\zs}{\mathbf{z}_{\mathrm{Style}}}

\newcommand{\X}{\mathbf{X}}
\newcommand{\Z}{\mathbf{Z}}

\newcommand{\D}{ {\cal D} }

\newcommand{\name}{VidStyleODE }

\renewcommand{\paragraph}[1]{{\vspace{1mm}\noindent \bf #1}.}

\usepackage{xr-hyper}

\makeatletter
\newcommand*{\addFileDependency}[1]{
  \typeout{(#1)}
  \@addtofilelist{#1}
  \IfFileExists{#1}{}{\typeout{No file #1.}}
}

\makeatother

\usepackage[pagebackref,breaklinks,colorlinks]{hyperref}
\usepackage[capitalize]{cleveref}
\crefname{section}{Sec.}{Secs.}
\Crefname{section}{Section}{Sections}
\Crefname{table}{Table}{Tables}
\crefname{table}{Tab.}{Tabs.}
\crefname{figure}{Fig.}{Figs.}

\usepackage{lipsum}
\usepackage{multirow}
\usepackage{booktabs}
\usepackage{tabu}
\usepackage[accsupp]{axessibility}  %
\begin{document}

\title{\paperTitle}

\author{ \authorBlock
\and
Tolga Birdal\\
Imperial College London\\
{\tt\small tbirdal@imperial.ac.uk}
\and
Duygu Ceylan\\
Adobe Research\\
{\tt\small ceylan@adobe.com}
\and
Levent Karacan\\
Iskenderun Technical University\\
{\tt\small levent.karacan@iste.edu.tr}
\and
Erkut Erdem\\
Hacettepe University\\
{\tt\small erkut@cs.hacettepe.edu.tr}
\and
Aykut Erdem\\
Koç University\\
{\tt\small aerdem@ku.edu.tr}
}

\twocolumn[{
\renewcommand\twocolumn[1][]{#1}%
\maketitle
\vspace{-0.4in}
\begin{center}
\vspace{-10pt}
    \centering
    \includegraphics[width=\textwidth]{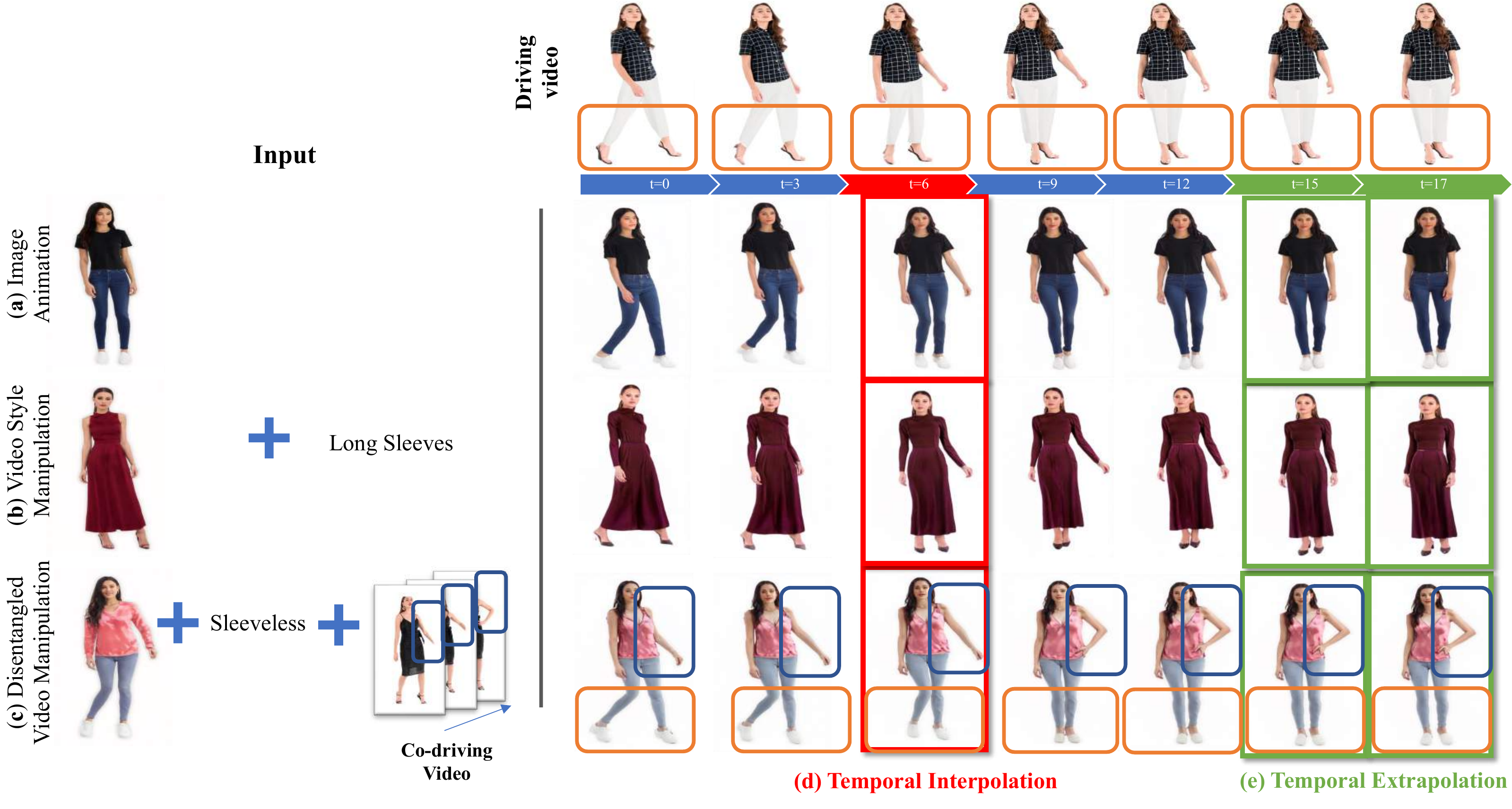}
    \captionof{figure}{\textbf{\name}~provides a spatiotemporal video representation in which motion and content info are disentangled, making it ideal for: (a) animating images, (b) consistent video appearance manipulation based on text, (c) body part motion transfer (\blue{blue} boxes) from a co-driving video while preserving remaining driving video dynamics (\orange{orange} boxes) intact, (d) temporal interpolation, and (e) extrapolation. \emph{Zoom in for better viewing.}
    }
    \label{fig:teaser}
\end{center}%
}]
\maketitle

\setlength{\textfloatsep}{0.5cm}

\begin{abstract}\vspace{-3mm}
We propose \textbf{\name}, a spatiotemporally continuous disentangled \textbf{vid}eo representation based upon \textbf{Style}GAN and Neural-\textbf{ODE}s. Effective traversal of the latent space learned by Generative Adversarial Networks (GANs) has been the basis for recent breakthroughs in image editing. However, the applicability of such advancements to the video domain has been hindered by the difficulty of representing and controlling videos in the latent space of GANs. In particular, videos are composed of content (i.e., appearance) and complex motion components that require a special mechanism to disentangle and control. To achieve this, {\name} encodes the video content in a pre-trained StyleGAN $\mathcal{W}_+$ space and benefits from a latent ODE component to summarize the spatiotemporal dynamics of the input video. Our novel continuous video generation process then combines the two to generate high-quality and temporally consistent videos with varying frame rates. We show that our proposed method enables a variety of applications on real videos: text-guided appearance manipulation, motion manipulation, image animation, and video interpolation and extrapolation. Project website: 
\textcolor{blue}{\href{https://cyberiada.github.io/VidStyleODE/}{https://cyberiada.github.io/VidStyleODE}}
\end{abstract}

\section{Introduction}
\label{sec:intro}
Semantic image editing is revolutionizing the visual design industry by enabling users to perform accurate edits in a fast and intuitive manner. Arguably, this is achieved by carrying out the \emph{image manipulation} process with the guidance of a variety of inputs, including text~\cite{Li_2020_CVPR,tedigan,Bau-PaintByWord-2021,styleclip}, audio~\cite{Lee_2022_CVPR,Li_2022_ECCV}, or scene graphs~\cite{Dhamo_2020_CVPR}. 
Meanwhile, the visual characteristics of real scenes are constantly changing over time due to various sources of motion, such as articulation, deformation, or movement of the observer. Hence, it is desirable to adapt the capabilities of image editing to videos. Yet, training generative models for high-res videos is challenging due to the lack of large-scale, high-res video datasets and the limited capacity of current generative models (\eg GANs) to process complex domains. This is why the recent attempts \cite{stylefacev, videogpt} are limited to low-res videos.
Approaches that treat videos as a discrete sequence of frames and utilize image-based methods (\eg \cite{Karacan_2022_BMVC, latenttransformer, stitch}) also suffer from important limitations such as a lack of temporal coherency and cross-sequence generalization.

To overcome these limitations, we set out to learn \textbf{spatio-temporal} video representations suitable for both generation and manipulation with the aim of providing several desirable properties. First, representations should \textbf{express} \textbf{high-res} videos accurately, even when trained on low-scale low-resolution datasets. Second, representations should be robust to \textbf{irregular} motion patterns such as velocity variations or local differences in dynamics, \ie deformations of %
articulated objects. Third, it should naturally allow for \textbf{control and manipulation of appearance and motion}, where manipulating one does not harm the other
\eg manipulating motion should not affect the face identity. 
We further desire to learn these representations \textbf{efficiently} on extremely sparse videos (3-5 frames) of arbitrary lengths. 
To this end, we introduce {\name}, a principled approach that learns disentangled, spatio-temporal, and continuous motion-content representations, which possesses all the above attractive properties. 

Similar to recent successful works~
\cite{latenttransformer,alaluf2022times,stitch,Karacan_2022_BMVC}, we regard an input video as a composition of a fixed appearance, often referred to as video \emph{content}, with a motion component capturing the underlying \emph{dynamics}.
Respecting the nature of \emph{editing}, we propose to model latent \emph{changes} (\emph{residuals}) required for taking the source image or video towards a target video, specified by an external \emph{style} input \emph{and/or} co-driving videos. For this purpose,  {\name}~first disentangles the content and dynamics of the input video. We model content as a global code in the $\Wplus$ space of a \emph{pre-trained} StyleGAN generator and regard dynamics as a continuous signal encoded by a latent ordinary differential equation (ODE) ~\cite{Rubanova2019,NODE,ballas2015delving}, ensuring temporal smoothness in the latent space. \name~then explains all the video frames in the latent space as \emph{offsets} from the single global code summarizing the video content. These offsets are computed by solving the latent ODE until the desired timestamp, followed by subsequent self- and cross-attention operations interacting with the dynamics, content, and style code specified by the textual guidance. To achieve effective training, we omit adversarial training that is commonly used in the literature and instead introduce a novel temporal consistency loss (Sec. \ref{sec:network}) based on CLIP \cite{clip}. We show that it surpasses conventional consistency objectives and exhibits higher training stability.

Overall, our contributions are:
\begin{enumerate}[noitemsep,topsep=0pt,leftmargin=*]
    \item We build a novel framework, \name, disentangling content, style, and motion representations using StyleGAN2 and latent ODEs.
    \item By using latent directions with respect to a global latent code instead of per-frame codes, \name~ enables external conditioning, such as text, leading to a simpler and more interpretable approach to manipulating videos.
    \item We introduce a new \emph{non-adversarial} video consistency loss that outperforms prior consistency losses, which mostly employ conv3D features, at a lower training cost. 
    \item We demonstrate that despite being trained on low-resolution videos, our representation permits a wide range of applications on high-resolution videos, including appearance manipulation, motion transfer, image animation, video interpolation, and extrapolation (\cf~\cref{fig:teaser}).
\end{enumerate}

\section{Related Work}
\label{sec:related}
\paragraph{GANs} Since their introduction, GANs \cite{gan, stylegan2} have achieved great success in synthesizing photorealistic images. %
Recent methods \cite{psp, e4e, pti} obtain the latent codes of real images in 
 StyleGAN's latent space and modify them to achieve guided manipulation considering the task at hand \cite{tedigan, styleclip, hairclip}. Despite their ability to generate high-res images, GANs are deemed challenging to train on complex distributions such as full-body images \cite{stylehuman, insetgan} or videos. Earlier attempts \cite{mocogan, saito_2020, temporal_gan, styleganv} modified GAN architecture to effectively synthesize videos based on sampled content and motion codes. Most notably, StyleGAN-V \cite{styleganv} recently modified StyleGAN2 to synthesize long videos while requiring a similar training cost. However, these methods are bounded by the resolution of the training data and are impractical for complex domains and motion patterns. Our work leverages the expressiveness of a pre-trained StyleGAN2 generator to encode input videos as trajectories in the latent space and extends image-based editing strategies to enable consistent text-guided video appearance manipulation.

\paragraph{Video generation} %
Recent works focused on using a pre-trained image generation as a video generation backbone. MoCoGAN-HD \cite{mocogan-hd} and StyleVideoGAN~\cite{fox2021stylevideogan} synthesize videos from an autoregressively sampled sequence of latent codes. InMoDeGAN \cite{InMoDeGAN} decomposes the latent space into semantic linear sub-spaces to form a motion dictionary. Other methods \cite{stylefacev, Video2StyleGAN} decompose pose from identity in the latent space of pre-trained StyleGAN3, enabling talking-head animation from a driving video. StyleHeat \cite{StyleHEAT} warps intermediate pre-trained StyleGAN2 features with predicted flow fields for video/audio-driven reenactment. \cite{MRAA, wang2022latent} animate images based on a driving video following optical-flow-based methods in the pixel \cite{MRAA} or latent \cite{wang2022latent} space. Despite their success, these methods are limited to unconditional video synthesis \cite{mocogan-hd, fox2021stylevideogan}, are restricted to a single domain  \cite{stylefacev, Video2StyleGAN, StyleHEAT}, designed for a single purpose \cite{MRAA, wang2022latent, StyleHEAT, stylefacev, Video2StyleGAN}, and/or incapable to effectively generate high-res videos \cite{styleganv}. We present a domain-invariant framework to learn disentangled representations of content and motion, enabling a range of applications on high-res videos. In contrast to all of the aforementioned methods except MRAA \cite{MRAA}, we also do not use adversarial training. %
With the motivation of handling irregularly sampled frames and continuous-time video generation, some previous works also incorporated latent ODEs \cite{NODE} for unconditional video generation \cite{vidode}, future prediction from single frame \cite{simpleode}, or modeling uncertainty in videos \cite{vae2ode}. Despite being limited to low-res videos, these methods showed the potential of latent ODEs in video interpolation and extrapolation. \name~further extends them by showing the effectiveness of latent ODEs in high-res video interpolation and extrapolation.

\paragraph{Semantic video manipulation}
Applying image-level editing to individual video frames often leads to temporal incoherence. To alleviate this problem, Latent Transformer \cite{latenttransformer} uses a shared latent mapper to the latent codes of the input frames in a pre-trained StyleGAN2 latent space. Alaluf et al.~\cite{alaluf2022times} propose a consistent video inversion/editing pipeline for StyleGAN3. STIT \cite{stitch} fine-tunes a StyleGAN2 generator on the input video and moves along a single latent direction to realize the target edit. These methods still fail to achieve temporally consistent manipulation due to the entanglement between appearance and video dynamics in the StyleGAN space, defying their presumption of temporal independence between video frames. As a remedy, DiCoMoGAN \cite{Karacan_2022_BMVC} encodes video dynamics with a neural ODE \cite{NODE}, and learns a generator that manipulates input frames based on the learned motion dynamics and a target textual description. StyleGAN-V \cite{styleganv} enables video manipulation by projecting real videos onto a learned content and motion space, enabling appearance manipulation via the modification of the content code following image-based methods \cite{hairclip, styleclip}. Instead of directly modifying  content code, our model achieves guided manipulation by discovering spatio-temporal latent directions conditioned on the target description and the video dynamics. This allows for greater flexibility regarding the appearance-motion entanglement of StyleGAN space. \name~also encodes video dynamics with a latent ODE that encourages a smooth latent trajectory, thus enhancing temporal consistency.

\begin{figure*}
    \centering
    \includegraphics[width=\textwidth]{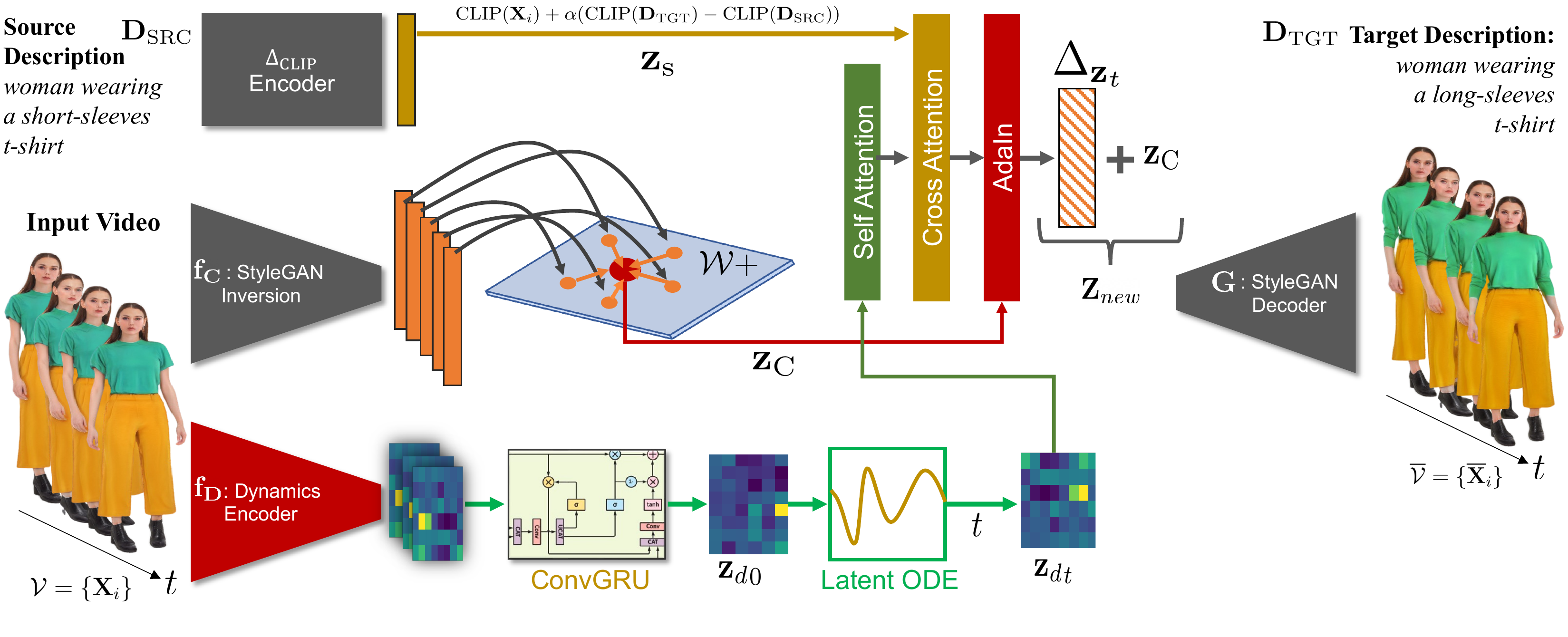}
    \caption{\textbf{\name~overview}. We encode video dynamics and process them using a ConvGRU layer to obtain a dynamic latent representation $\Z_{d0}$ used to initialize a latent ODE of the motion (bottom). We also encode the video in $\Wplus$ space to obtain a global latent code $Z_C$ (middle). We combine the two with an external style cue through an attention mechanism to condition the AdaIN layer that predicts the directions to the latent codes of the frames in the target video (top). Modules in \textcolor{gray}{\textbf{gray}} are \emph{pre-trained} and \emph{frozen} during training.
    \vspace{-4mm}}
    \label{fig:model_diagram}
\end{figure*}

\section{Method}
\label{sec:method}

We consider an input video $\Vid=\{\X_i\in\R^{M\times N\times 3}\}_{i=1}^K$ consisting of $K$ RGB frames along with an associated textual description $\Dsrc$. Our goal is to explain $\Vid$ by learning an explicitly manipulable \emph{continuous representation} conditioned on an external \emph{style} input. As manipulation is inherently related to making \emph{changes}~\cite{hairclip}, \name~achieves this goal via a deep neural architecture, modeling the changes through disentangled \emph{content}\footnote{set of attributes fixed along the temporal dimension~\cite{Karacan_2022_BMVC,mocogan, mocogan-hd}},  \emph{style}\footnote{attributes of interest subject to change} and \emph{dynamics}\footnote{an intrinsic force producing change}.
To this end, \name~first uses a pre-trained {spacetime encoder} $f_C:\Vid\to\z_C$ to summarize the information content of the input video frames or individual images as a \emph{global latent code}. 
Our key idea is to explain individual video frames with respect to the global code as \emph{translations} along the latent dimensions of a pre-trained high-res {image generator} $G(\cdot)$:
\begin{equation}
    \overline{\X}_t =  G\left( \z_{new} = \z_C + {\dz}_t \right)
\end{equation}
To find these \emph{latent directions} ${\dz}_t$ that entangle dynamics and style, we (i) continuously model latent representation of dynamics ${\z_d}_t$, which can be queried at arbitrary timesteps; (ii) learn to predict these directions by interacting with the global code $\z_C$ and the predicted dynamics ${\z_d}_t$, conditioned on the target style $\z_S$, while preserving the content. There are multiple ways to get $\z_S$, but in this work, we choose to extract it based on target and source textual descriptions $(\Dsrc,\Dtgt)$.
We first describe the method design for each of these components, depicted in~\cref{fig:model_diagram}, followed by implementation and architectural details in~\cref{sec:network}.

\paragraph{Spatiotemporal encoding $f_C$} 
To encode the entire video into a global code, we seek a \emph{permutation-invariant} representation of the input video, factoring out the temporal information. To this end, we first project all the frames in $\Vid$ onto the $\Wplus$ space of StyleGAN2~ \cite{stylegan2} by using an \emph{inversion}~\cite{xia2022gan} to obtain a set of \emph{local} latent codes $\Z:=\{\z^l_i\in\Wplus\}_{i=1}^K$. We then apply a symmetric pooling function to obtain the order-free global video content code: $\z_C = \Expect\left[\Z\right]$. %

\paragraph{Continuous dynamics representation}
Inspired by%
\cite{rempe2020caspr,park2021vid}, to model the spatiotemporal input, \ie, to compute representations for unobserved timesteps at arbitrary spacetime resolutions, we opt for learning a latent subspace $\z_{d0} \in\R^D$, that is used to initialize an autonomous latent ODE $\frac{d {\z_d}_t}{dt} = f_{\theta}({\z_d}_t)$, which can be advected in the latent space rather than physical space:
\begin{equation}
\label{eq:ode}
    {\z_d}_T = \phi_T({\z_d}_0) = {\z_d}_0 + \int_{0}^T f_{\theta}({\z_d}_t, t) \, dt
\end{equation}
where $\theta$ denotes the learnable parameters of the model $f_{\theta}$.
This (1) enables \emph{learning} a space best suited to modeling the dynamics of the observed data and (2) improves scalability due to the fixed feature size. Due to the time-independence of $f_{\theta}$, advecting ${\z_d}_{t=0}$ forward in time by solving this ODE until $t=T\geq 1$ yields a representation that can explain latent variations in video content. To learn the initial code ${\z_d}_0$, we encode each frame individually by a \emph{spatial encoder} $f_D:\X_i\to \R^{m_d\times n_d \times 64}$. Resulting tensors are fed into a ConvGRU$:\R^{m_d\times n_d \times 64\times K}\to\R^{m_{\mathrm{ode}}\times n_{\mathrm{ode}} \times 512}$~\cite{park2021vid,ballas2015delving} in reverse order so that the final code seen by the model corresponds to the first frame. %

The use of a Neural ODE here provides several benefits over other approaches such as an LSTM (see \cref{tab:ablation_table_2}). One especially important benefit is the ability to handle irregularly sampled frames during training, which allows for scaling to longer videos while keeping memory costs constant. Additionally, the ODE allows for extrapolation into unseen timesteps, due to this irregular training. Finally, Neural ODEs are able to better learn the geometry of the dynamic latent space, providing a meaningful space due to the powerful regularization that ODEs impose.

\begin{figure}[b]
    \centering
    \includegraphics[width=\columnwidth]{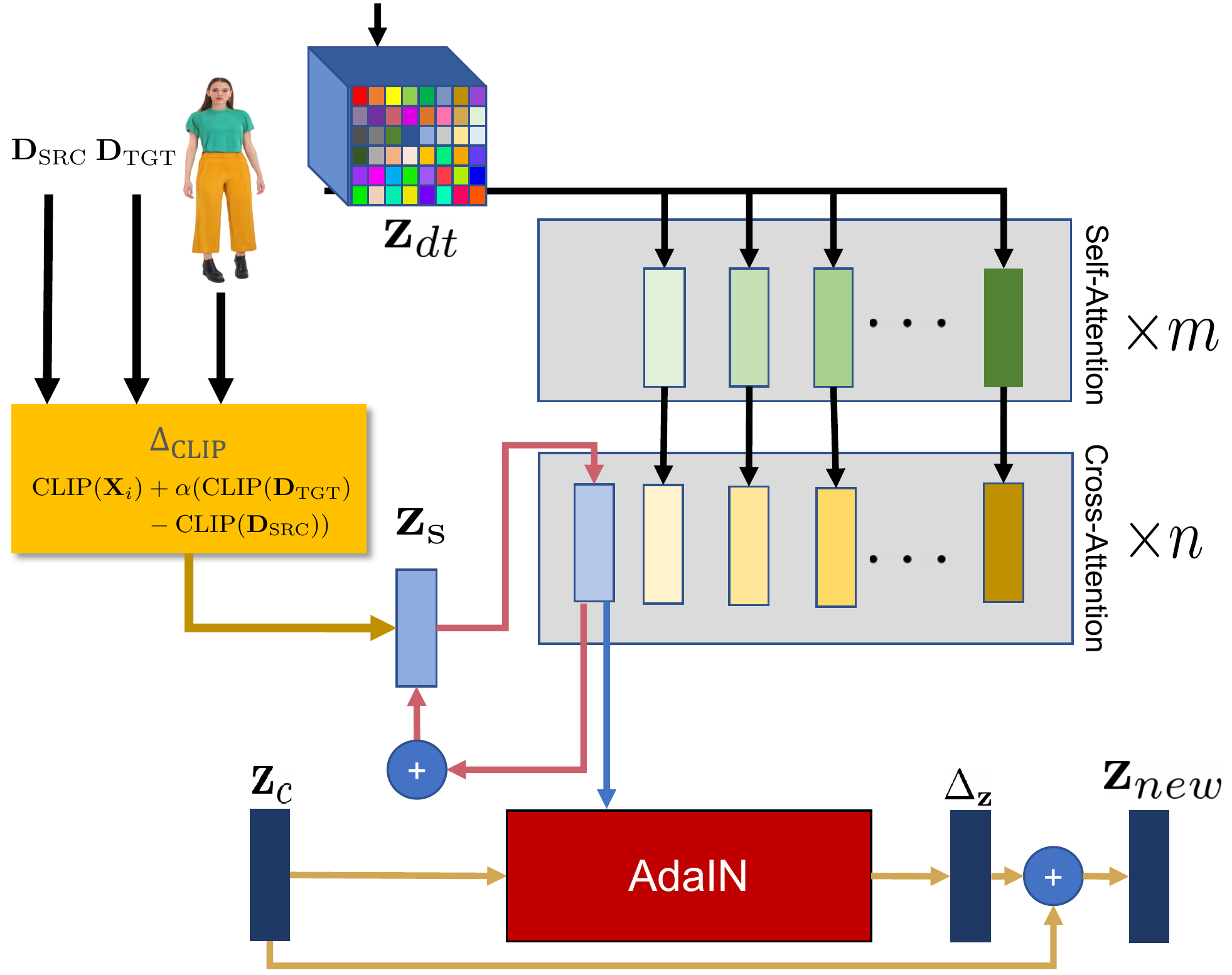}
    \caption{{\textbf{Proposed attention scheme utilized in VidStyleODE.}}
    }
    \label{fig:attention_diagram}
\end{figure}

\paragraph{Conditional generative model $f_G$} 
As illustrated in \cref{fig:attention_diagram}, to synthesize high-quality video frames that adhere to the target style $\z_S$, \name~generatively models the desired output at time $t$ as an explicit function of {content}, {dynamics} and {style}:
\begin{equation}
    \overline{\X}_t = G\left( \z_t \right), \quad \z_t = \fmerge(\z_c, \z_d\,|\, \z_S)= \z_C + {\dz}_t,
\end{equation}
where the \emph{latent direction} ${\dz}_t$ depicts the residual required to realize the desired edits and is computed by a series of self-attention (SA)~\cite{transformer}, cross-attention (CA)~\cite{transformer} and adaptive instance normalization (AdaIN)~\cite{adain} operators:
\begin{equation}
    {\dz}_t = \mathrm{AdaIn}\left(\mathrm{CA}(\mathrm{SA}({\z_d}_t), \z_S), \z_C\right)
\end{equation}
Modeling the \emph{change} in this manner rather than the target latents themselves is significantly less complex and allows for manipulating the given video in relation to its global code. As such, and as we demonstrate experimentally, it offers significant advantages of fidelity and manipulation-ability. We implement $G(\cdot)$ as a pre-trained StyleGAN2 generator.

\paragraph{Obtaining the text-driven style $\z_S$}
We model the \emph{change} in source and target descriptions as a \emph{style direction} ${\dz}^{\mathrm{Style}}=\CLIP(\Dtgt) - \CLIP(\Dsrc)$
in the CLIP latent space~\cite{clip,styleclip}. We then move towards this direction in the CLIP space to obtain the text conditioning code: 
\begin{equation}\label{eq:clip}
\z_S = \CLIP(\X_i) + \alpha \dz^{\mathrm{Style}}%
\end{equation}
where $\alpha$ is a user-controllable parameter determining the scale of the manipulation.

\subsection{Training and Network Architectures}
\label{sec:network}
We train \name~by minimizing a multi-task loss $\loss$ over the text-video pairs to find the best parameters of dynamics encoder $\fmot$ as well as $\fmerge$ while keeping the content encoder and the image generator frozen:
\begin{equation}
    \loss = \lambda_{C} \loss_{C} + \lambda_{A} \loss_{A} \\ + \lambda_{S} \loss_{S} + \lambda_{D} \loss_{D} + \lambda_{L} \loss_{L}
\end{equation}
where $\lambda_{*}$ depicts the corresponding regularization coefficients. We next detail each of these terms, which are consistency, appearance reconstruction, structure reconstruction, CLIP directional loss, and latent direction regularization.

\paragraph{CLIP consistency loss}
DietNeRF \cite{dietnerf} shows that the CLIP \cite{clip} image similarity score is more sensitive to changes in appearance, compared to those caused by varying viewpoints. This led the authors to propose a new consistency loss as the pair-wise CLIP dissimilarity between images rendered from different viewpoints in order to guide the reconstruction of 3D NeRF representation. We observe that CLIP is also more sensitive to changes in appearance than to changes in dynamics Thus, we propose to replace the expensive temporal discriminator used in the literature \cite{mocogan, vid2vid, mocogan-hd}, with a CLIP consistency loss along the temporal dimension. Specifically, we sample $N_C$ frames from the generated video and minimize the pair-wise dissimilarity between them. %
\begin{equation}
    \loss_{C}(\Vid) = \sum_{i=1}^{N_C} \sum_{j \geq i}^{N_C} 1 - (\CLIP_I(\overline{\X}_{i})^{T} \CLIP_I(\overline{\X}_{j}))
\end{equation}
where $\overline{\X}_i$ is the $i_{th}$ sampled frame from the generated video, and $\CLIP_I$ is the CLIP image encoder. %

\paragraph{Appearance and structure reconstruction loss}
To learn the video dynamics, previous work \cite{Karacan_2022_BMVC, stylefacev, StyleHEAT, Video2StyleGAN} commonly used a VGG perceptual loss and L2 loss, which reconstructs both the structure and appearance of the input video. This inherently requires the image generator to be fine-tuned on the input video dataset. Considering that most available video datasets are of a low resolution and low diversity, fine-tuning the image generator on these datasets would greatly affect the model's capability to generate diverse and high-quality videos. Therefore, we propose to use a disentangled structure/appearance reconstruction loss to guide learning the dynamic representation. In particular, we employ the Splicing-ViT \cite{Splice2022} appearance loss to encourage the appearance of the generated video to match the appearance represented in the global code $\z_C$. Additionally, as motion dynamics are closely related to the change in structure \cite{structure}, we utilize Splicing-ViT structural loss to encourage the dynamics of the generated video to follow the dynamics of the input video. 
\begin{eqnarray}
       \loss_{A} &= \sum_{i=1}^{N} ||ViT_A(G(\z_C)) - ViT_A(G(\z_{t_i}))|| \\
    \loss_{S} &= \sum_{i=1}^{N} ||ViT_S(\X_i) - ViT_S(G(\z_{t_i}))|| 
\end{eqnarray}
where $ViT_A$, and $ViT_S$ are the latent features in DINO-ViT \cite{dinovit} corresponding to appearance and structure, respectively, as described in \cite{Splice2022}. 
This way, we can disentangle learning appearance and dynamic representation completely, enabling diverse high-res video generation via low-res video datasets.

\paragraph{CLIP video directional loss}
Given source and target descriptions, and a reference image, \cite{gal2021stylegannada} proposes to guide the appearance manipulation in the generated image by encouraging the change of the images in the CLIP space to be in the same direction as the change in descriptions. We adapted this loss to the video domain using:
\begin{eqnarray}
    \Delta_T &= \CLIP_T(T_{desc}) - \CLIP_T(S_{desc}) \\\nonumber
    \Delta_V &= \frac{\sum_{1}^{N}{\CLIP_I(\overline{\X}_i) - \CLIP_I(G(\z_{t_i}))}}{N} \\
    \loss_{D} &= 1 - {\Delta_V \Delta_T}\,/\,{|\Delta_V| |\Delta_T|}\nonumber
\end{eqnarray}
 where $\CLIP_T$, and $\CLIP_I$ correspond to the CLIP text and image encoder, respectively, and $N$ refers to the number of sampled frames from the generated video. During training, we sample three frames per video.
 
\paragraph{Latent direction loss}
We regularize the norm of the latent directions ${\dz}$ to prevent the model from following directions with large magnitudes: $\loss_{L} = \Expect [||{\dz}_{t_i}||]_i$. We observed that this loss also helped in making the model converge faster.

\paragraph{Network architectures}
We used a ResNet architecture adapted from \cite{park2020swapping} as our dynamic encoder $f_D$. Additionally, we used Vid-ODE ConvGRU network \cite{park2020vidode} to obtain the dynamic representation $\z_d$ before utilizing the Dopri5 \cite{torchdiffeq} method to solve the first-order ODE. We apply self-attention and cross-attention over $\z_d$ by dividing the input tensor into patches and treating them as separate tokens, following \cite{vit-dosovitskiy2020image}. Additionally, we used a pSp encoder to obtain $\z_i$, and a StyleGAN2 generator \cite{stylegan2} for $G(\cdot)$, pre-trained on Stylish-Humans-HQ Dataset \cite{stylehuman} for fashion video experiments, and on FFHQ \cite{karras2019style} for face video experiments.

\paragraph{Training details}
Thanks to our choice of modeling dynamics as a latent ODE, we are able to train on irregularly sampled frames. Specifically, for every training step, we sample $k$ different frames from each input video and a target description from other videos in the batch. We use those to compute the aforementioned losses. Details about hyperparameters can be found in the supp. materials. %

\section{Experimental Analysis}
\label{sec:results}

\paragraph{Datasets and prepossessing.}
We evaluated our method mainly on the recent dataset of Fashion Videos \cite{Karacan_2022_BMVC} composed of $3178$ videos of fashion models and RAVDESS dataset \cite{ravdess}, containing $2,452$ videos of $24$ different actors speaking with different facial expressions. We split each dataset randomly into 80\% train and 20\% test data. Moreover, we aligned their video frames following \cite{stylehuman, stylegan2}, and downsampled the input videos during training to \textbf{$128 \times 96$} for Fashion  $128 \times 128$ for RAVDESS. Additionally, we annotated each actor in RAVDESS according to gender, hairstyle, hair color, and eye color, and procedurally generated target descriptions based on these attributes. 

\paragraph{Evaluation metrics}
To assess the performance of the models, we use the following metrics. \emph{Frechet Video Distance} (FVD) \cite{fvd} measures the difference in the distribution between ground truth (GT) videos and generated ones. \emph{Inception Score} (IS) \cite{inception_score} and \emph{Frechet Inception Distance} (FID) \cite{frechet_inception_distance} measures the diversity and perceptual quality of the generated frames. 
\emph{Manipulation Accuracy} quantifies the agreement of the edited video with the target text, relative to a GT video description.
\emph{Warping error}\cite{warp_error} measures the temporal appearance consistency. \emph{Average key-point distance} (AKD) assesses the structural similarity  between the generated and driving videos. \emph{Average Euclidean distance} (AED) evaluates identity preservation in reconstructed videos. 

\paragraph{Baselines}
We compare our method against the state-of-the-art text-guided video manipulation and image animation approaches, namely Latent Transformer (LT) \cite{latenttransformer}, DiCoMoGAN \cite{Karacan_2022_BMVC}, STIT \cite{stitch}, StyleGAN-V \cite{styleganv}, and MRAA \cite{MRAA}. As LT requires separate training for each target attribute, we trained it to manipulate only the sleeve length on Fashion Videos and averaged its performance for RAVDESS on gender, hair, and eye color. Additionally, we trained DiCoMoGAN and StyleGAN-V on the face and fashion datasets using the same alignment process in our method. STIT fine-tunes the generator using PTI \cite{pti} for each input video, taking ~10 minutes for a 1-minute video on NVIDIA RTX 2080, and further uses image-based manipulation methods.  We employed StyleCLIP global directions. StyleGAN-V achieves text-guided manipulation by performing test-time optimization of projected latent codes with CLIP. We also considered HairCLIP \cite{hairclip} and StyleCLIP \cite{styleclip} as baselines for frame-by-frame manipulation of the video. Lastly, we train MRAA \cite{MRAA} and adapt StyleGAN-V code to evaluate same-identity and cross-identity image animation. (\cf supplementary materials).%

\begin{figure}[!t]
    \centering
    \includegraphics[width=\columnwidth]{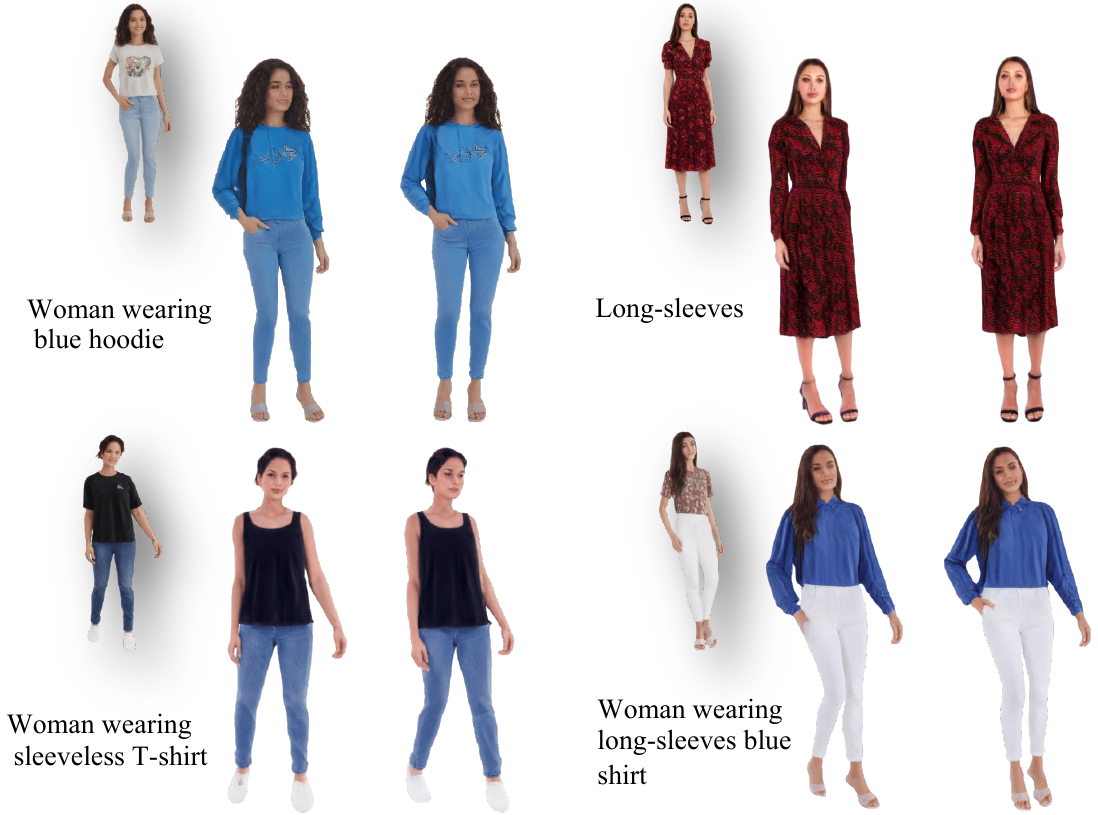}
    \caption{\textbf{Text-guided editing results}. \name~ lets the users manipulate a frame based on a text prompt, and transfer manipulated attributes to other videos in a consistent way. Source frames are shown at the top left corner along with the target texts. %
    }
    \label{fig:manipulation}
\end{figure} 
\subsection{Results}

\begin{figure}[!t]
    \centering
    \includegraphics[width=0.99\linewidth, height=10.4cm]{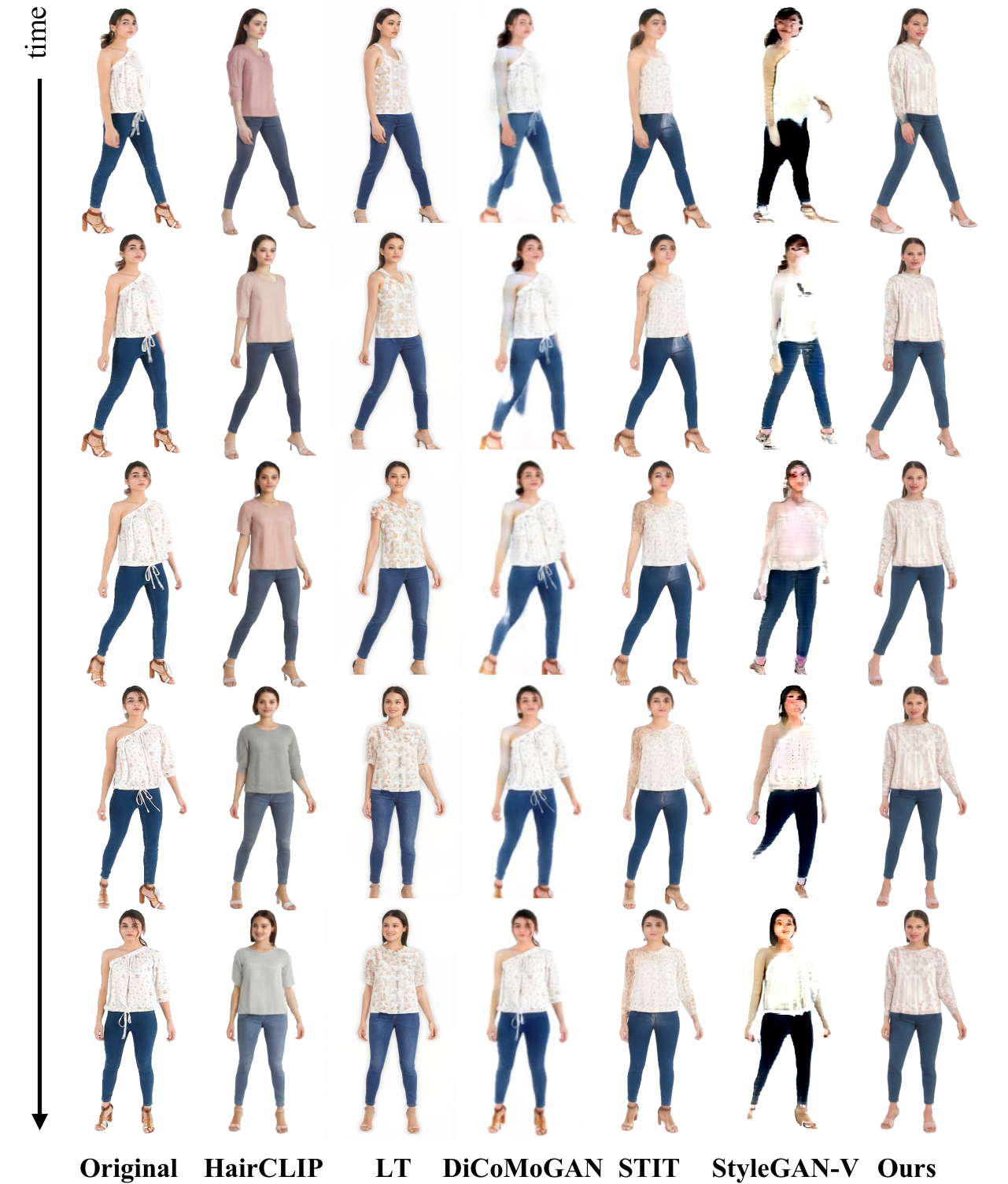}
    \caption{\textbf{Qualitative comparison against the state-of-the-art}. \name~produces more realistic results than existing semantic video methods when changing sleeve length from short to long, with improved visual quality and manipulation accuracy. HairCLIP, a frame-level method, lacks temporal coherence.
    }
    \label{fig:baselines}
\end{figure}

\paragraph{Semantic video editing}
Our method allows for text-guided video editing by conditioning the prediction of the latent direction on the manipulation direction specified by the target and source descriptions. \cref{fig:manipulation} shows that our method accurately manipulates the color, clothing style, and sleeve length in a temporally-consistent way on several sample video frames. \name~can also handle target descriptions that consider either single or multiple attributes without introducing artifacts. \cref{fig:baselines} compares our method against the state-of-the-art. As seen, LT~\cite{latenttransformer} and the frame-level HairCLIP~\cite{hairclip} fail to preserve temporal consistency, especially with respect to the identity. DiCoMoGAN~\cite{Karacan_2022_BMVC} and STIT~\cite{stitch} perform poorly in applying meaningful and consistent manipulations. In particular, DiCoMoGAN fails to perform the necessary manipulations in the text-relevant parts such as the sleeves, and produces artifacts in the text-irrelevant parts. STIT applies the same latent direction to all of the video frames in the StyleGAN2 $\mathcal{W}_+$ space. We show that this is prohibitive, as the relative edits of the manipulated parts, such as the sleeves' length, change as the body~moves. %

These observations are also reflected in the results reported in~\cref{tab:metrics_table}.
As LT cannot jointly manipulate multiple attributes with the same model, we consider a relatively simple setup where we only manipulate the length of the sleeves of the source garments for a fair comparison.
STIT, which performs instance-level optimization, gives the best FVD, yet its manipulation accuracy is significantly inferior to ours. 
Although HairCLIP achieves the best accuracy metric, its performance is the worst in terms of (temporal) video quality as measured by FVD. Our \name~method achieves an FVD close to STIT, and a manipulation accuracy close to HairCLIP. In general, it is the only method that produce smooth and temporally-consistent videos with high fidelity to the target attributes. It also preserves the identity of the person while making the target garment edits.

\begin{table*}[!t]
\centering
\setlength{\tabcolsep}{2mm} 
\resizebox{0.99\textwidth}{!}{

\begin{tabular}{p{2.6cm}@{$\qquad\;$}ccccc@{$\qquad\;$}ccccc}
    \toprule
    
\multirow{2}{*}{Method} & \multicolumn{5}{c}{\textbf{Fashion Videos}} & \multicolumn{5}{c}{\textbf{RAVDESS}}\\
    & FVD $\downarrow$ & IS $\uparrow$ & FID $\downarrow$ & Acc. $\uparrow$ & $W_{error}$ $\downarrow$  & FVD $\downarrow$ & IS $\uparrow$ & FID $\downarrow$ & Acc. $\uparrow$ & $W_{error}$ $\downarrow$  \\
    \midrule
    
HairCLIP~\cite{hairclip} & 548.09 & 2.56 & 65.57 & \textbf{0.92} & 0.0152 \vspace{1mm} & \underline{218.70} & 1.33 & \underline{31.47} & \underline{0.83} & 0.0136 \\ 
STIT~\cite{stitch} & \textbf{126.04} & 3.08 & \underline{33.24} & 0.72 & 0.0089 & 226.31 & 1.33 & 32.89 & 0.71 & 0.0088 \\ %
LT~\cite{latenttransformer} & 262.17 & \underline{3.08} & 39.06 & 0.24 & 0.0095 & 339.48 & \underline{1.35} & 37.05 & 0.43 & 0.0192\\ %
DiCoMoGAN~\cite{Karacan_2022_BMVC} & 324.30 & 2.50 & 103.62 & 0.51 & 0.0151 & \textbf{121.92} & \textbf{1.40} & \textbf{16.38} & 0.38  &  \underline{0.0086} \\
StyleGAN-V~\cite{styleganv} & 988.96 & 2.30 & 135.49 & 0.71 & 0.0384 & 487.91 & 1.28 & 66.89 & \textbf{0.87} & 0.0307 \\
Ours & \underline{157.48} & \textbf{3.25} & \textbf{26.28} & \underline{0.87} & \textbf{0.0075} & 273.10 & 1.33 & 34.92 & \underline{0.83} &  \textbf{0.0076}\\
\bottomrule
\end{tabular}
}
\caption{\textbf{Quantitative comparison on the Fashion and RAVDESS datasets}. We report the performances using metrics for evaluating photorealism (FVD, IS, and FID), manipulation accuracy (Acc.), and temporal coherency ($W_{error}$). While the scores in \textbf{bold} highlight the best performance, the \underline{underlined} ones show the second best. Overall, our \name~method is the only approach that gives photorealistic and temporally consistent results with accurate edits of the garment attributes.
} 
\label{tab:metrics_table}
\end{table*}

\begin{figure}[!t]
    \centering
    \includegraphics[width=0.995\columnwidth]{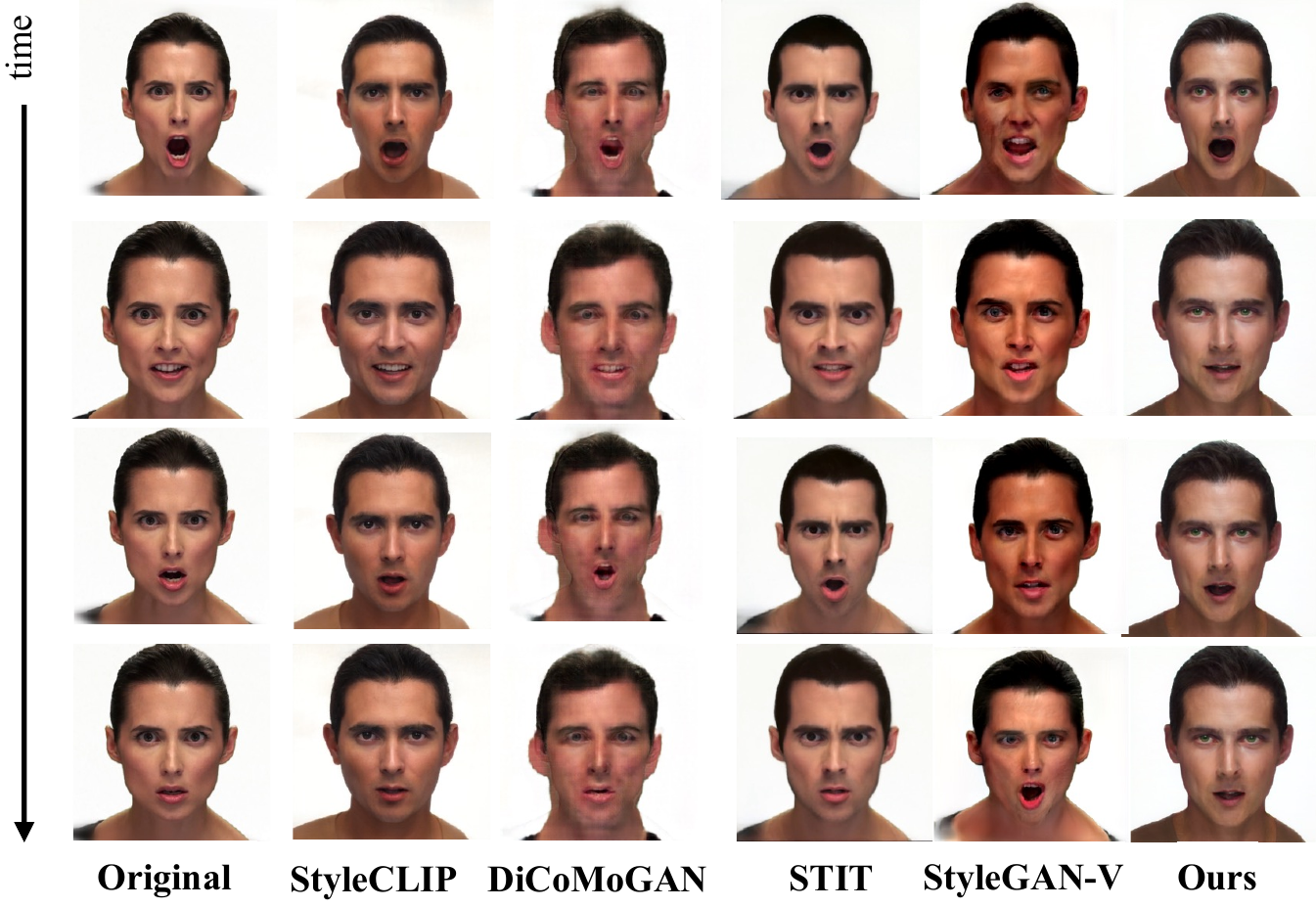}
    \caption{\textbf{Facial attribute manipulation}. Target Description: a photo of a man with \emph{green eyes}. \name~gives a temporally consistent output when manipulating source face video, unlike other methods which show inconsistencies in hairline, nose, or identity, or fails to make the proper edits.
    }
    \label{fig:baselines_face}
\end{figure}

\cref{fig:baselines_face} shows further manipulation results on the RAVDESS dataset. We observe that existing models exhibit similar limitations observed in the Fashion Videos dataset but at a lower degree. We hypothesize that this is mainly due to StyleGAN2 learning a more disentangled and expressive latent space on a simple dataset containing face images.

In summary, we conclude that auto-encoder-based approaches such as \cite{Karacan_2022_BMVC} are able to faithfully reconstruct the text-irrelevant parts such as the face identity but lack the capability of performing meaningful manipulations, resulting in artifacts and unnatural-looking videos. StyleGAN2-based approaches \cite{hairclip, stitch} achieve good semantic manipulation but lack the ability to keep a consistent appearance in the generated video. \name~benefits from a pre-trained StyleGAN2 generator to perform meaningful semantic manipulations while producing smooth and consistent videos. 

\begin{figure}[!t]
    \centering
    \includegraphics[width=\columnwidth]{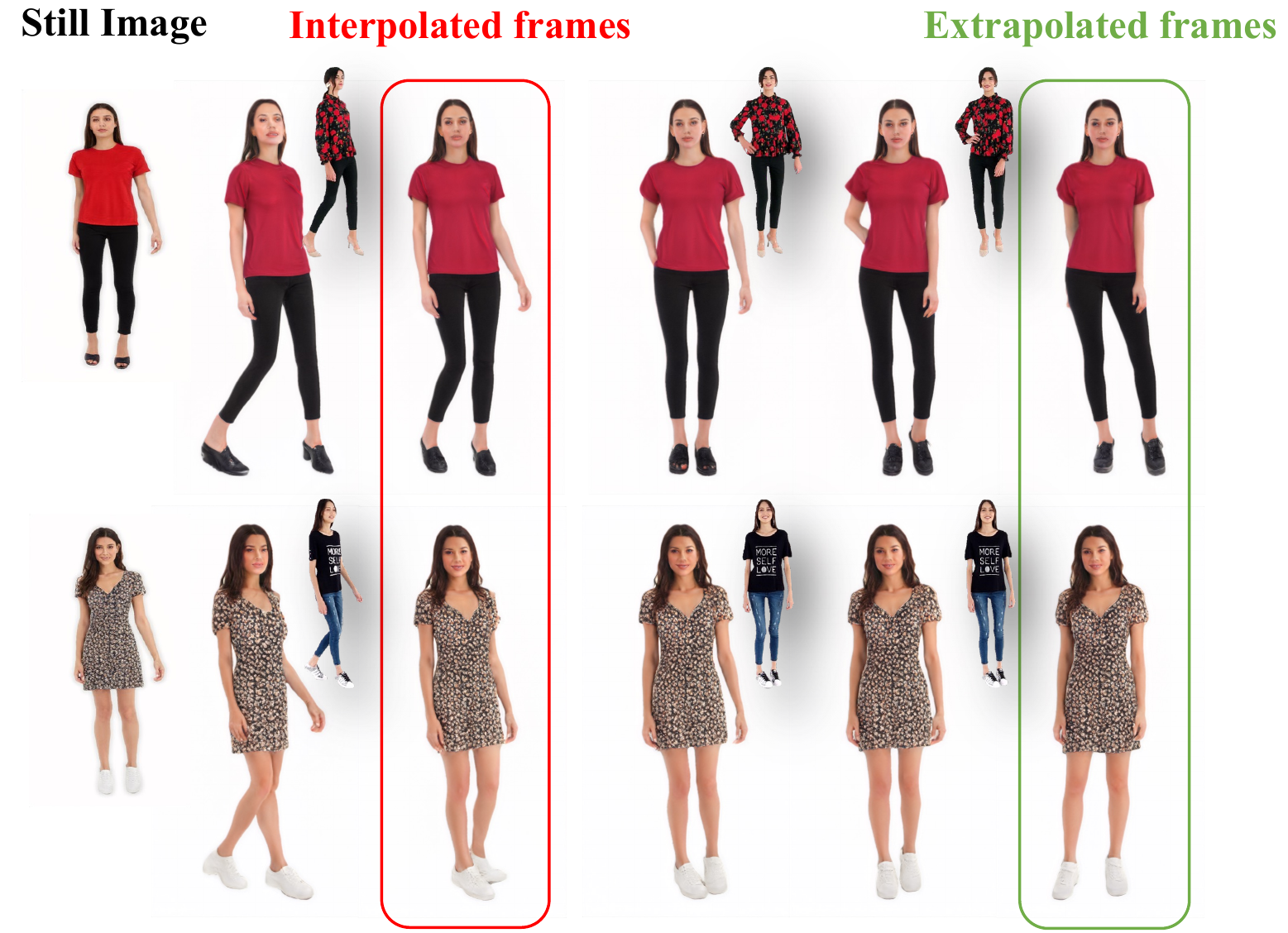}
    \caption{\textbf{Animating a still image}. Our method animates input images using motion dynamics from a driving video. With a learned continuous representation of motion dynamics via a latent ODE, it can also generate realistic frames via interpolation or extrapolation.
    }
    \label{fig:animation}
\end{figure}

\begin{figure}[!t]
    \centering
    \includegraphics[width=0.99\columnwidth]{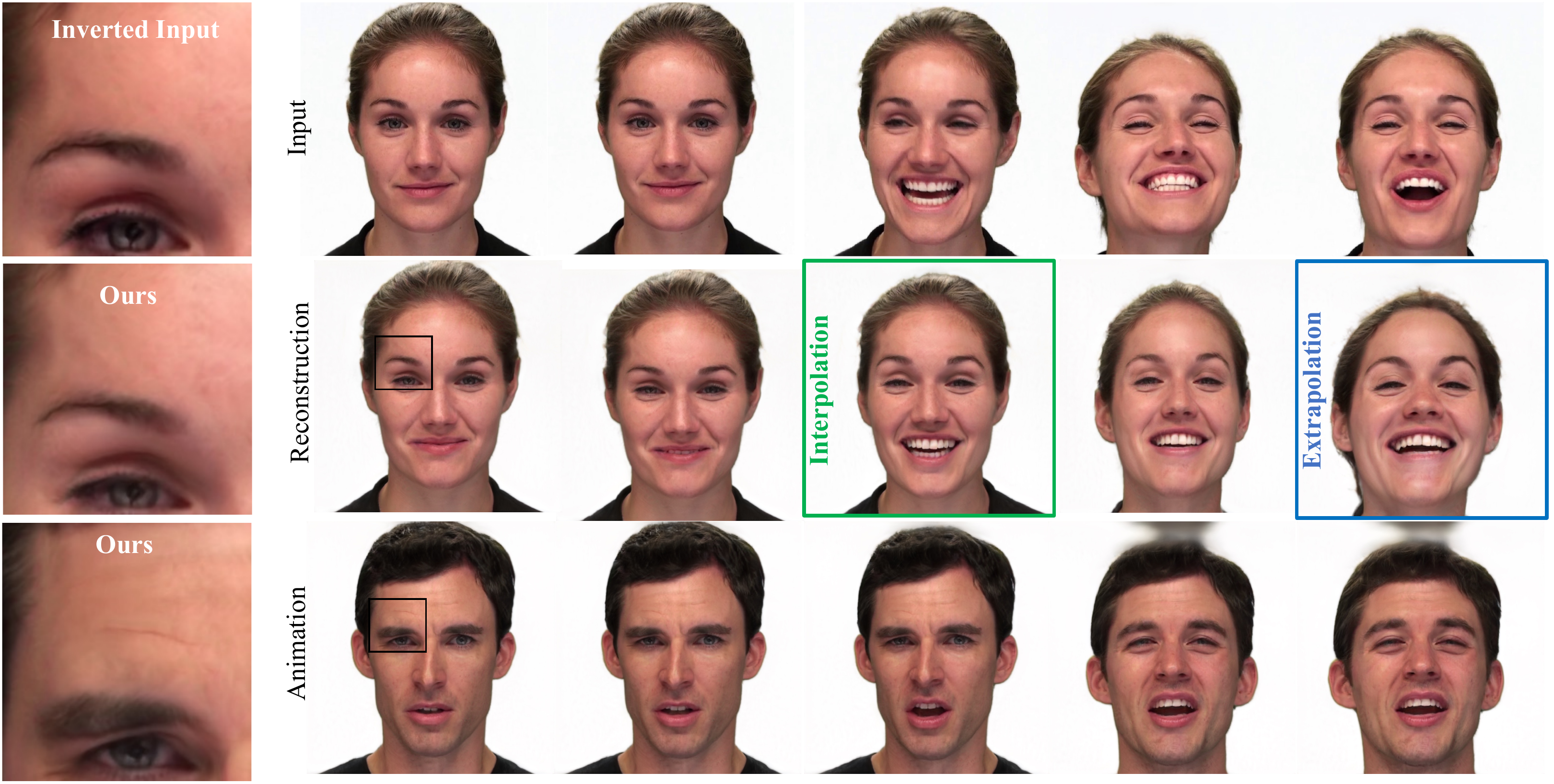}
    \caption{\textbf{High-resolution results on RAVDESS.} \name~ maintains the perceptual quality of the pre-trained and frozen StyleGAN2 Generator (col. 1), while enabling temporal interpolation (col. 4) and extrapolation (col. 6), and image animation (last row).
    }
    \label{fig:perceptual}
\end{figure}

\begin{figure}[!t]
    \centering
    \includegraphics[width=0.99\columnwidth]{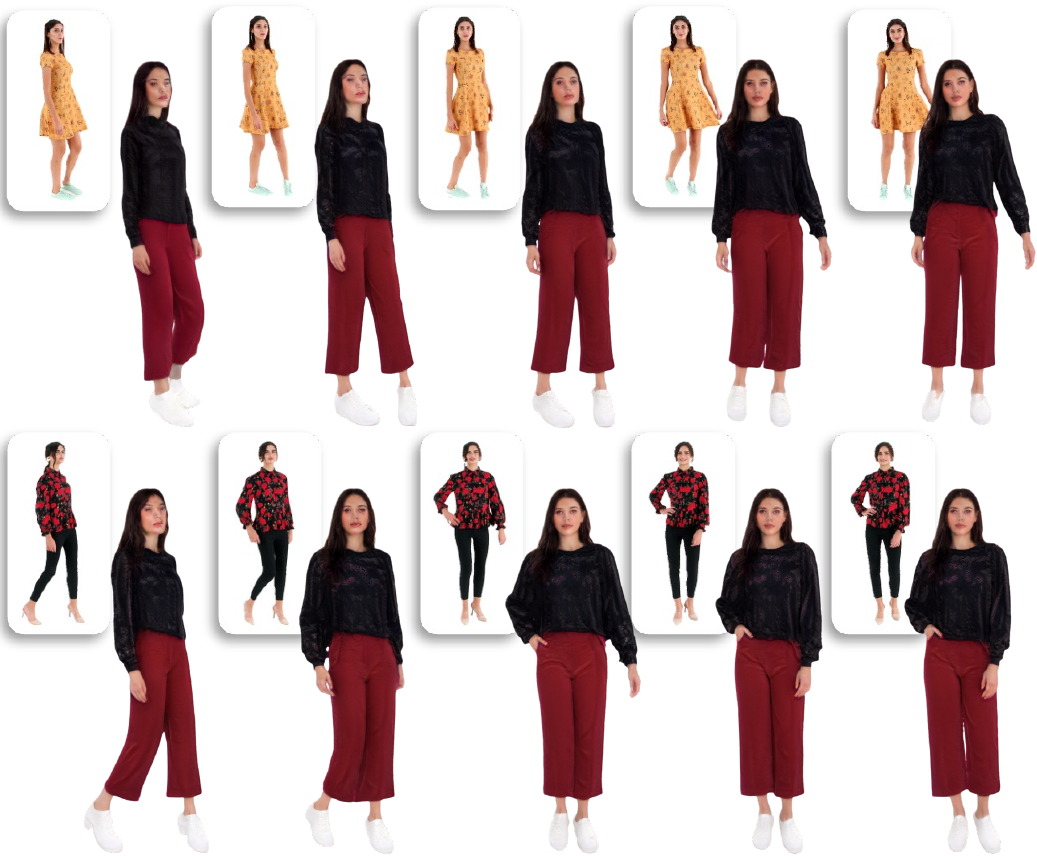}
    \caption{\textbf{Diverse animation results achieved by \name}. Each example shows a separate driving video (top-left corner) and the corresponding animations. Our method provides disentangled motion control while keeping the source content information intact.
    }
    \label{fig:manipulation_w_disentanglement}
\end{figure}

\paragraph{Image animation and video interpolation/extrapolation}
Our model is able to learn a disentangled representation of content and motion, allowing for animating the content extracted from a still image using the motion dynamics coming from a driving video. In \cref{fig:animation} and \cref{fig:perceptual}, we show some sample results of this process. Since our framework is equipped with a latent ODE, we can use our method to perform interpolation between selected video frames. 
Moreover, we are able to extrapolate the motion dynamics to future timesteps not seen in the original driving video. \cref{fig:manipulation_w_disentanglement} further shows the ability of our method in controlling the motion dynamics in a disentangled manner. As seen, we can obtain diverse animations of a given source image by transferring motion from different driving videos. Our method generates a consistent appearance for the person across different videos (Table~\ref{tab:metrics_table2}).

\begin{table}[!t]
\centering
\setlength{\tabcolsep}{1mm} 
\resizebox{\columnwidth}{!}{
\begin{tabular}{p{2.6cm}ccc@{$\quad$}ccc}
    \toprule
    
\multirow{2}{*}{Method} & \multicolumn{3}{c}{\textbf{Fashion Videos}} & \multicolumn{3}{c}{\textbf{RAVDESS}}\\
  & AKD$_C$ $\downarrow$ &  AKD$_S$ $\downarrow$ & AED$_S$ $\downarrow$  & AKD$_C$ $\downarrow$ &  AKD$_S$ $\downarrow$ & AED$_S$ $\downarrow$  \\
\midrule
    
StyleGAN-V~\cite{styleganv} & 12.76 & 10.24  & 0.29 & 3.36 & 2.17 & 0.16 \\
MRAA~\cite{MRAA} & 10.67 & \textbf{2.46} & 0.25 & \textbf{2.65} & \textbf{1.08} & \textbf{0.12} \\
Ours & \textbf{6.15} & 5.46 & \textbf{0.22} & 2.86 & 2.12 & 0.16 \\
\bottomrule
\end{tabular}}
\caption{Quantitative comparison on cross-identity (C) and same-identity (S) image animation. Our method achieves competitive results to SOTA image animation approaches as a byproduct of encoding video dynamics with Latent-ODEs.%
} 
\label{tab:metrics_table2}
\end{table}

\begin{figure}[!t]
    \includegraphics[width=\columnwidth]{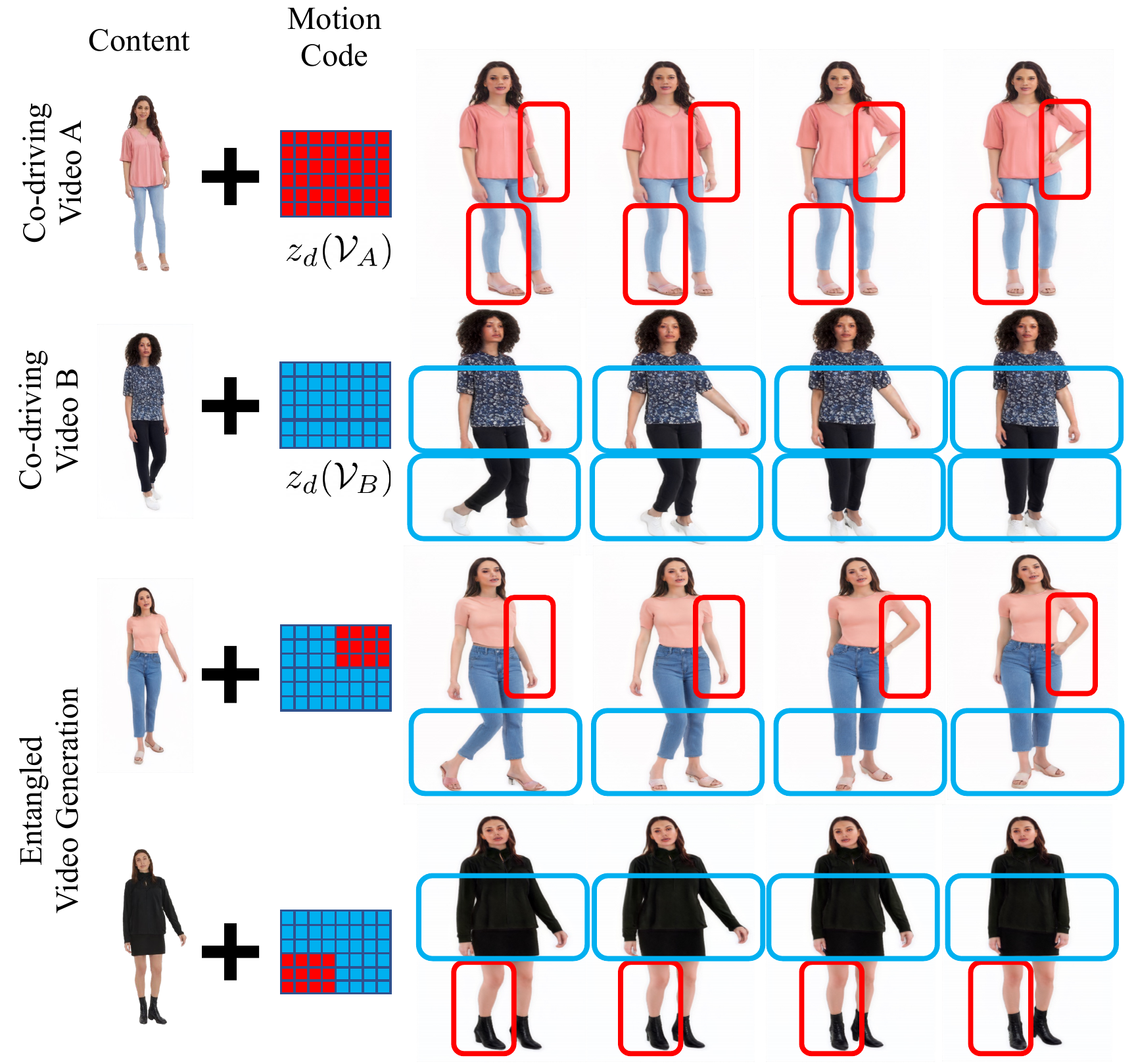}
    \caption{\textbf{Local motion dynamics control}. 
    \name~can blend motion from two co-driving videos $A$ and $B$, whose dynamics are depicted in first two rows. %
    The last two rows show \name's ability to transfer dynamics from these driving videos in a local manner. The \red{red} and the \blue{blue} boxes encode spatial regions where the motion dynamics are extracted and transferred.%
    }
    \label{fig:dyn_local_manipulation}
\end{figure}

\paragraph{Controlling local motion dynamics.}
We observed a local correspondence between \name~dynamic latent representation and video motion dynamics, allowing for transferring local motion of body parts between different videos. In particular, given $\z_{d_A}\in\R^{8\times 8}$ and $\z_{d_B}\in\R^{8\times 8}$ corresponding to videos $A$ and $B$ respectively, we follow a blending operation to obtain a new dynamic latent code $\z_{d_{new}}$ as $\z_{d_{new}} = m \z_{d_A} + (1 - m) \z_{d_B}$ where $m \in \{ 0, 1 \} ^ {8 \times 8}$ is a spatial mask. In \cref{fig:dyn_local_manipulation}, we show an example of transferring different body part movements (right hand or left leg) from different videos. To the best of our knowledge, we are the first that manage to control local motion dynamics.
Additional results can be found in the supplementary.

\paragraph{Ablation study}
\label{par:ablation}
We split the ablation into two parts, focusing on different aspects of our approach. 

\cref{tab:ablation_table} shows the contribution of each loss to the overall performance where we remove each one at a time and report how the metrics are affected. Omitting the CLIP consistency loss ${L}_{C}$ causes an increase in both warping error and the FVD score. Replacing the CLIP consistency loss with either a StyleGAN-V or MoCoGAN-HD temporal discriminator also leads to a worse performance in both metrics. Moreover, eliminating the prediction of latent residuals ${\dz}_t$ and instead computing the final vector $\textbf{z}_{t}$ directly causes a considerable drop in the FVD score. Replacing the appearance loss ${L}_{A}$ and structure loss $L_{S}$ with a VGG perceptual loss produces more temporally inconsistent video. %

Moreover, \cref{tab:ablation_table_2} focuses on evaluating the components of our approach. In particular, we test replacing the Neural ODE with an LSTM, removing the self-attention layer entirely, and replacing the cross-attention layer with a concatenation of between $\z_c$, $\z_S$, and the output of the self-attention layer. We observe that both the self- and cross-attention layers are essential for the realism of the video, as indicated by the relatively worse FVD and IS scores.
Moreover, replacing the ODE with a two-layer LSTM leads to a significant drop in the performance across all metrics. We also found that the LSTM-based approach results in an $\approx 74\%$ increase in training time and restricts the number of frames during training to 30 frames on a single V100, as opposed to the irregular sampling in the ODE which allows for handling longer videos.

\begin{table}[!t]
\centering
\setlength{\tabcolsep}{6mm} 
\resizebox{\columnwidth}{!}{
\noindent\begin{tabular}{lcc}
    \toprule
\hspace{-2mm}Model Details & FVD $\downarrow$  & $W_{error}$ $\downarrow$ \\
    \hline
\hspace{-1mm}\name & 157.48  & 0.0075 \\
\hspace{-1mm}$\quad$ w/o $\mathcal{L}_{C}$ & 191.08 &  0.0095\\
\hspace{-1mm}$\quad$ w/o $\mathcal{L}_{C}$, w/ SD & 229.87 & 0.0084\\
\hspace{-1mm}$\quad$ w/o $\mathcal{L}_{C}$, w/ MD & 245.04 & 0.0115\\
\hspace{-1mm}$\quad$ w/o $\mathcal{L}_{C}$, latent residuals & 222.76 & 0.0097 \\
\hspace{-1mm}$\quad$ w/o $\mathcal{L}_{A}$, $\mathcal{L}_{S}$, and $\mathcal{L}_{C}$ & 244.49  &  0.0125 \\
    \bottomrule
\end{tabular}
}
\caption{\textbf{Ablation analysis of losses on Fashion Videos}. MD refers to the temporal discriminator introduced in MoCoGAN-HD\cite{mocogan-hd} and SD refers to the temporal discriminator from StyleGAN-V \cite{styleganv}.}
\label{tab:ablation_table}
\end{table}

\section{Conclusion}
\label{sec:conclusion}
We have presented VidStyleODE, a novel method to disentangle the content and motion of a video by modeling \emph{changes} in the StyleGAN latent space. To the best of our knowledge, it is the first method using a Neural ODE to represent motion in conjunction with StyleGAN, leading to a well-formed latent space for dynamics. By modifying content-dynamics combinations in different ways, we enable various applications. We have also introduced a novel consistency loss using CLIP that improves the temporal consistency without requiring adversarial training. %

\paragraph{Limitations \& future work}
While we freeze the pre-trained StyleGAN generator to prevent any perceptual quality degradation, it may lead to an identity shift in the generated videos and less consistent appearance due to the limited expressiveness of the generator. Fine-tuning the generator and the inversion network on the video dataset can reduce this problem as discussed in the supplementary materials. Albeit omitted, a future work may benefit from task-driven \emph{test-time training} to resolve the aforementioned problems without affecting the perceptual quality. Additionally, we noticed an over-smoothed motion on the datasets with periodic motion, such as RAVDESS. This is a limitation of autonomous first-order ODEs, which struggle with forming closed-loop solutions on periodic dynamics due to the uniqueness theorem. Future work may employ higher-order ODEs to enhance the dynamics representation on such datasets. Moreover, we invite the community to explore text-guided editing of local dynamics in the future.

\begin{table}[!t]
\centering
\renewcommand{\arraystretch}{1.2}
\fontsize{15pt}{15pt}\selectfont
\setlength{\tabcolsep}{6mm} 
\resizebox{\columnwidth}{!}{
\noindent\begin{tabular}{l@{$\;\;$}c@{$\;\;$}c@{$\;\;$}c@{$\;\;$}c@{$\;\;$}c}
    \toprule
\hspace{-2mm}Model Details & FVD $\downarrow$ & IS $\uparrow$ & Acc. $\uparrow$ & $W_{error}$ $\downarrow$ & $AKD_{S}$ $\downarrow$\\
    \hline
\hspace{-1mm}\name & 157.48  & 3.25 & 0.87 & 0.0075 & 5.03 \\
\hspace{-1mm}$\;\;\;$ w/o ODE (w/ LSTM) & 350.95 & 2.81 & 0.81 & 0.0095 & 6.00 \\
\hspace{-1mm}$\;\;\;$ w/o Self-Attn & 256.30 & 2.80 & 0.98 & 0.0067 & 5.21 \\
\hspace{-1mm}$\;\;\;$ w/o Cross-Attn (w/ Concat) & 240.21 & 2.89 & 0.96 & 0.0068 & 5.33 \\

    \bottomrule
\end{tabular}
}
\caption{\textbf{Ablation of different model components on Fashion Videos.} Removing the self-attention or cross-attention layers yields substantially worse FVD and IS scores, while providing only minor improvements in other metrics. Additionally, replacing the ODE component with an LSTM yield worse performance across all metrics.}
\label{tab:ablation_table_2}%
\end{table}

{
\section*{Acknowledgements} The authors would like to thank KUIS AI Center for letting them use their High-Performance Computing Cluster. Tolga Birdal wants to thank Google for their gifts. 

\bibliographystyle{ieee_fullname}
\bibliography{11_references}
}

\ifarxiv \clearpage \onecolumn

\maketitle

In this supplementary document, we discuss several design choices, introduce ablation studies, and implementation details. We also provide additional qualitative and quantitative results both on Fashion Videos and RAVDESS datasets. To view a comprehensive collection of videos from all our different applications, you can access the website \textcolor{blue}{\href{https://cyberiada.github.io/VidStyleODE/}{https://cyberiada.github.io/VidStyleODE}}
\vspace{4mm}

{
  \hypersetup{linkcolor=black}
  \tableofcontents
}
\vspace{5mm}

\section{Discussions}
We now discuss some interesting trends, comparisons, and trade-offs between models, supported by quantitative and qualitative results. 

\paragraph{On inversion vs. quality}
HairCLIP~\cite{hairclip} manages to consistently obtain high results on manipulation accuracy across both datasets. However, it is highly dependent on high-quality inversions to obtain good results. On the Fashion dataset, where the base inversion quality is not good enough, we see very bad results across the other metrics. However, on RAVDESS, the inversion quality is much higher, due to taking advantage of the models trained on FFHQ. Therefore, we see very good results across all metrics.

\paragraph{Perceptual quality vs. manipulation capability}
STIT~\cite{stitch} performs quite well on consistency and perceptual quality metrics, primarily due to the fine-tuning of the generator. However, by ensuring high-quality results, it reduces the ability to manipulate the videos effectively, and so is consistently behind multiple other models. Additionally, high-frequency details of the videos (such as shoes, hair, and complex color patterns) are lost due to the focus on reducing distortion. 

DiCoMoGAN~\cite{Karacan_2022_BMVC} primarily acts as an autoencoder, with additional steps for manipulation. On RAVDESS, where the videos are not too complicated to learn, this allows DiCoMoGAN to obtain very high results on all the perceptual quality metrics. However, this auto-encoding property also restricts the ability to manipulate accurately, leading to poor results for that metric.

\paragraph{Complexity of dynamics vs. generation quality}
StyleGAN-V~\cite{styleganv} had a lot of challenges learning the correct motion of the Fashion dataset and frequently suffered from mode collapse. This led to very poor perceptual quality results, as well as a very high warping error. On the RAVDESS dataset, there were fewer training issues, which contributed to relatively better results. However, all the perceptual quality metrics are still very poor.

The primary issue with MRAA~\cite{MRAA} is the inability to distinguish between what motion should be transferred and what should not, as well as retaining key structural details of the reference frame. As seen in~\cref{fig:driving1}, MRAA transforms the dress into pants, following the style of the driving video. Additionally, because the sleeve of the right arm is not visible in the reference frame, it attempts to copy the sleeve style of the driving video, leading to inconsistencies between the two sleeve lengths. In~\cref{fig:driving2}, it also attempts to transfer sleeve length, even when the right arm is visible in the reference frame. Not but not least, MRAA also removes a lot of the fine detail of the clothing. Therefore, despite being able to properly capture the motion of the people, in both cases, it is unable to create a complete and consistent video.

\paragraph{On trade-offs}
Notably, most models were unable to handle all tasks effectively: \textbf{generation}, \textbf{disentanglement}, and \textbf{manipulation}. While some are very good at manipulation, others obtained high perceptual quality. However, \name~hits a sweet spot. On the Fashion dataset, it consistently achieves very good results across all metrics, including being the best in many of them. On RAVDESS, \name~is able to achieve very good consistency and manipulation accuracy, while still reporting competitive perceptual quality metrics. Therefore, it does not suffer from the same trade-offs between manipulation, consistency, and perceptual quality as the other models.

\section{Architectural Details}
\paragraph{Spatiotemporal encoder $f_C$} We use a pre-trained StyleGAN2 inversion network to obtain the $K$ input frames' latent representation in the $\Wplus$ space $\Z:=\{\z^l_i\in\Wplus\}_{i=1}^K$. We freeze the inversion network's weights during training. Then, we take the expectation of $\Z$ to obtain the video's global latent code $\z_C$. During inference, the global latent code can be sampled or obtained from a single frame. In our experiments, we used pSp inversion network \cite{psp} pre-trained on StylishHumans-HQ Dataset \cite{stylehuman} for fashion video experiments, and
on FFHQ \cite{karras2019style} for face video experiments.

\paragraph{Dynamic representation network $f_D$} We first process the $K$ video frames $X_i \in \R^{M \times N \times 3}$ independently using a $2D$ ResNet encoder architecture based on the implementation of \cite{park2020swapping} to extract $K$ feature maps $\z_r \in \R^{m_d \times n_d \times d_{sp}}$. In our experiments with the fashion videos dataset, we used $M = 128$,  $N = 96$, $m_d=8$, $n_d=6$, and $d_{sp} = 64$. Additionally, for face videos experiments, we used $M = 128$,  $N = 1128$, $m_d=8$, $n_d=8$, and $d_{sp} = 64$. Subsequently, we adapt ConvGRU from \cite{park2020vidode} to extract dynamic latent representation $\z_d \in \R ^ {m_{ode} \times n_{ode} \times 512}$ from $\z_R =\{\z_{r_i}\}_{i=1}^K$. For all of our experiments, we set $m_{ode} = m_{d}$ and $n_{ode}=n_{d}$. We use the dynamic representation to initialize an autonomous latent ODE 
\begin{equation}
\label{eq:ode_app}
    {\z_d}_T = \phi_T({\z_d}_0) = {\z_d}_0 + \int_{0}^T f_{\theta}({\z_d}_t, t) \, dt,
\end{equation}
Where $z_{d0} = z_d$. We parameterize $f_\theta$ as a convolutional network obtained from \cite{park2020vidode}. For every training batch, we sample $n$ frames from each video and solve the ODE at their corresponding timestamps to obtain their spatiotemporal feature representation $\z_{dT} =\{\z_{d_{t_i}}\}_{i=1}^n$.  

\paragraph{Obtaining style code.} To guide the manipulation, we condition the video reconstruction on an external style code $\zs$. We represent this style code in the CLIP \cite{clip} embedding space by encoding the content frame $X_c$, source description $\Dsrc$ of the appearance of the video, and a target description $\Dtgt$. To obtain the content frame, we decode the latent global code using a pre-trained StyleGAN2 generator $G(\cdot)$. 
\begin{equation}\label{eq:clip_supp}
\zs = \CLIP_I(G(\z_C)) + \alpha  (\CLIP_T(\Dtgt) - \CLIP_T(\Dsrc))%
\end{equation}
where $\CLIP_I$ and $\CLIP_T$ are the CLIP image and text encoder, respectively. $\alpha$ is a user-defined parameter that controls the level of manipulation during inference time. For all of our quantitative experiments, we used $\alpha=1$. 

\paragraph{Conditional generator model} Once the video global code $\z_c$, the frames dynamic representation $z_{d}$, and the video style $z_{Style}$ have been collected, we apply $N$ layers of self-attention onto the different spatial components of $z_{d}$. Then, we perform cross-attention between the outputs of the self-attention and the style vector $z_{Style}$. At each layer of cross-attention, we predict and apply an offset to the style code in the CLIP space. We then take the final output style vector and modulate it over the global code $z_{c}$. This produces our direction, which is then added to the original code:
\begin{equation}
    \z_{t} = \z_{c} + \Delta \z
\end{equation}
The output frame at time $t$ is then generated as 
\begin{equation}
    \X_t = G\left( \z_t \right)
\end{equation}

\paragraph{Hyper-parameters}
The appearance and structural losses both have $\lambda_{S} = 10, \lambda_{A} = 10$. The latent loss has $\lambda_{L} = 1.0$. For the directional clip loss, we have $\lambda_{D} = 2.0$. For the consistency loss, we use a scheduler to go from $0.01$ to $1$ over $40000$ steps. For the trade-off between the structural and appearance loss, we use $\lambda = 0.5$, so that both are equally important. 

In our self-attention network, we use $12$ layers, each with $8$ heads, as well as a hidden dimension size of $512$. Both the coarse and medium layers receive the dynamics, while the fine layers do not.

\section{Details on the Datasets \& Evaluations}

\paragraph{Datasets}
All our results were evaluated on the Fashion dataset \cite{Karacan_2022_BMVC} and the RAVDESS dataset \cite{ravdess}. The Fashion dataset contains descriptions already, which we used for our manipulations. On RAVDESS, we hand-crafted descriptions for each of the 24 actors, which we used during training and testing for manipulation. 

\paragraph{Evaluation metrics}
We evaluate our model in terms of perception, temporal smoothness, and editing consistency of the generated videos as well as the accuracy of the applied manipulation. The \noindent\textit{Frechet Video Distance (FVD)} \cite{fvd} score measures the difference in the distribution between ground truth (GT) videos and generated ones, evaluating both the motion and visual quality of the video. To compute the metric, we used 12 frames sampled at 10 frames per second. \textit{Inception Score (IS)} \cite{inception_score} measures the diversity and perceptual quality of the generated frames. To eliminate any gain in IS from the diversity resulting in inconsistency in the video frames, we use only a single frame from each generated video. %
\textit{Frechet Inception Distance (FID)} \cite{frechet_inception_distance} measures the difference in distribution between GT and generated videos. Similar to IS, we use only a single frame from each generated video to calculate FID. \textit{Warping Error} predicts subsequent frames of a video using an optical flow network, and compares this with the generated frames, to measure consistency. The network we used is \cite{flowformer}. \textit{Manipulation Accuracy} measures the accuracy of the manipulation in the generated video according to the target textual description, and relative to the GT description of the video. We used CLIP \cite{clip} as a zero-shot classifier for this task. 

\paragraph{Baselines}
We trained HairCLIP \cite{hairclip} on the Fashion Videos dataset by omitting the attribute preservation losses concerning face images. For Latent Transformer, we followed the authors' instructions and trained the classifier for 20 epochs and the models for 10 epochs each. For HairCLIP, Latent Transformer, and STIT \cite{stitch} on the Fashion dataset, we used the StyleGAN-Human \cite{stylehuman} pre-trained generator. For RAVDESS, we used the FFHQ pre-trained generator for all 3 models. For DiCoMoGAN \cite{Karacan_2022_BMVC}, we trained the official code until convergence. 

For MRAA \cite{MRAA} on the Fashion dataset, we followed the training procedure provided by the authors for the tai-chi dataset. On the RAVDESS dataset, we used the training procedure provided for the VoxCeleb dataset.

We trained StyleGAN-V 3 times on each dataset for 1 week, using 2 V100 gpus. We picked the best model according to FVD (fvd2048\_16f), and used this for all metric calculations and figures. On both datasets, we noticed that later iterations suffered from significant mode collapse. Therefore, we also picked the epoch with the best FVD.
For manipulation, we projected real videos using 1000 iterations.

\section{Further Quantitative Results}
\begin{table*}[t]
\centering
\setlength{\tabcolsep}{2mm} 
\resizebox{0.99\textwidth}{!}{
\begin{tabular}
{p{2.6cm}@{$\qquad\;$}ccccc@{$\qquad\;$}ccccc}
    \toprule
    
\multirow{2}{*}{Method} & \multicolumn{5}{c}{\textbf{Fashion Videos}} & \multicolumn{5}{c}{\textbf{RAVDESS}}\\
    & FVD $\downarrow$ & IS $\uparrow$ & FID $\downarrow$ & Acc. $\uparrow$ & $W_{error}$ $\downarrow$  & FVD $\downarrow$ & IS $\uparrow$ & FID $\downarrow$ & Acc. $\uparrow$ & $W_{error}$ $\downarrow$  \\
    \midrule
    
Ours & 157.48 & 3.25 & \textbf{26.28} & 0.87 & \textbf{0.0075} & 273.10 & \textbf{1.33} & \textbf{34.92} & \textbf{0.83} &  0.0076\\
Ours w/ FT & \textbf{139.69} & \textbf{3.27} & 31.69 & 0.87 &  0.0096  & \textbf{160.90} & 1.32 & 36.48 & 0.79 & \textbf{0.0049} \\
\bottomrule
\end{tabular}
}
\caption{\textbf{Effect of fine-tuning the pre-trained generator and inversion network.}
Fine-tuning StyleGAN-2 image generator and inversion network (Ours w/ FT) significantly improves the FVD score, with a minimum effect on the perceptual quality and manipulation capabilities of our model. 
} 
\label{tab:fine-tuned-netowrks}
\end{table*}

\subsection{Fine-tuning pre-trained networks}
A key motivation for this work is to develop a method that can generate and manipulate high-resolution videos (e.g. $1024 \times 512$) even when trained on low-resolution ones (e.g. $128 \times 96$ for Fashion Videos).
This impacted our choices for the architecture design and training objectives. For instance, fine-tuning the pre-trained image generator on the low-resolution training video dataset defies our original motivation to generate high-resolution videos. Additionally, while reconstruction loss between the generated and ground truth frames has been used in prior work \cite{stylefacev,Video2StyleGAN,fox2021stylevideogan} to reconstruct local dynamics, it often trades the lower distortion with the worse perceptual quality. However, in certain scenarios where a high-resolution training dataset is available, fine-tuning the generator and inversion networks is possible. We report in~\cref{tab:fine-tuned-netowrks} the performances of~\name~on Fashion Videos, where a generator and an inversion network pre-trained on Stylish-Humans-HQ Dataset \cite{stylehuman} were fine-tuned on Fashion Videos at a resolution of $1024 \times 512$ for $200k$ iterations. We also present results of \name~with a pre-trained generator and inversion networks trained on FFHQ~\cite{stylegan} and fine-tuned on RAVDESS~\cite{ravdess} at a resolution of $1024 \times 1024$ for $250k$ iterations. For both experiments, we trained \name~at a low resolution, i.e. $128 \times 96$ for Fashion Videos and $128 \times 128$ for RAVDESS. 

\subsection{Further Ablation Studies}
In~\cref{par:ablation} of the main paper, we analyzed the contribution of each component of our model to the final FVD and $W_{error}$ scores on the Fashion Videos, showing the superiority of our proposed CLIP temporal consistency loss $\loss_C$ over the MoCoGAN-HD temporal discriminator or the StyleGAN-V discriminator, as well as the validity of our architecture choices. We further analyze the effect of different strategies for obtaining the video global latent code. Specifically, we consider using the $\Wplus$ latent code of the first frame, a random frame, or the mean latent code.~\cref{tab:frame_number} shows that encoding the video content as the mean $\Wplus$ latent code (\ie $\z_C = \Expect\left[\Z\right]$) provides an overall better FVD, with less sensitivity to the frame order during inference (\eg First vs Last Frame).
\begin{table}
\begin{center}
\begin{tabular}{l|*{2}{c}}
\centering
\backslashbox[1cm]{Inference}{Training}
&{First Frame}&{Mean Frame}\\\toprule
First Frame & 169.62 & 181.64 \\\midrule
Random Frame & 206.98 & 182.39 \\\midrule
Mean Frame & 229.92 &  150.59\\\bottomrule
\end{tabular}
\end{center}
\caption{A comparison of different methods to obtain the global content representation for the Fashion dataset, in terms of FVD over same-identity image animation. Each row represents a different method of training, while each column represents inference using the stated global representation. Encoding video content with the mean $\Wplus$ latent code of the input frames during training provides a better FVD score with less sensitivity to the content frame position during inference.}
\label{tab:frame_number}
\end{table}

\section{Further Qualitative Results}
In this section, we provide additional qualitative results obtained with our proposed \name~and further comparisons against the state-of-the-art. 

\subsection{Latent motion representation}
Our model is able to learn a meaningful latent space for motion, which enables multiple applications. As seen in~\cref{fig:dynamic_interpolation}, interpolating between two motion representations produces a smooth combination of the two motions. Additionally, our motion representation contains spatial dimensions as well as a time dimension. By having access to local representations as well as a global representation for motion, we are able to manipulate only certain spatial parts of a video, or optionally the entire video.~\cref{fig:local_dyn_editing} shows the swapping of the upper right quarter of the motion representation while keeping the rest of it untouched. This is able to affect only the upper right quarter of the video, corresponding to the arm movement. Meanwhile, the other parts of the body follow the original motion path. We are also able to swap the global motion representation between two videos, resulting directly in the swap of the dynamics of the videos. This immediately allows for image animation, when combined with the global content representation, which can be obtained from a single frame if necessary. This is done by using the global motion representation from the driving video while extracting the global content representation from the source frame. Examples and comparisons with a popular image animation model \cite{MRAA} are found in~\cref{fig:driving1,fig:driving2}.

\subsection{Fashion Videos dataset}
\begin{itemize}[noitemsep,leftmargin=*,topsep=0pt]
    \item \cref{fig:dynamic_interpolation} shows the results of interpolating between two dynamic representations. Note especially the decrease in right arm movement, which happens smoothly as $\lambda$ moves from $0.0$ to $1.0$. Also, the legs spread out less at $t=25$ as we increase $\lambda$.
    \item In~\cref{fig:local_dyn_editing}, we provide an additional example of controlling local motion dynamics of a figure. In particular, we show that our model is able to transfer the right arm movements of another figure to the target figure.
    \item \cref{fig:baselines} provides qualitative comparisons for our \name model to the existing methods.
    \item \cref{fig:driving1} shows a comparison of our \name~model to multiple image animation methods. It can be seen that our method is the only one to transfer motion while maintaining the structure and perceptual quality of the reference frame. 
    \item \cref{fig:driving2} provides a second comparison of image animation methods. 
    \item \cref{fig:ablation} supports the quantitative ablations provided in the main paper with qualitative results, showing that the quality of the video does indeed improve as we add more components to the model.
    \item \cref{fig:video_manipulation,fig:video_manipulation_2} show text-guided editing examples on two sample videos from the Fashion Video dataset for two distinct target texts for each source video. As clearly seen, our method, \name, performs the necessary edits as suggested by the target descriptions successfully. 
    \item \cref{fig:interpolation} shows the ability of our method in generating realistic and consistent frames via interpolation. Our method accurately estimates the latent codes of the frames with the missing timestamps and generates the frames in an accurate manner.
    \item In~\cref{fig:extrapolation_1_frame}, we demonstrate that our \name~method animates still frames via extrapolation. As seen, it generates a video depicting visually plausible and temporally consistent movements, showing the effectiveness of our method.
\end{itemize}

\begin{figure*}[!t]
    \centering
    \includegraphics[width=\linewidth]{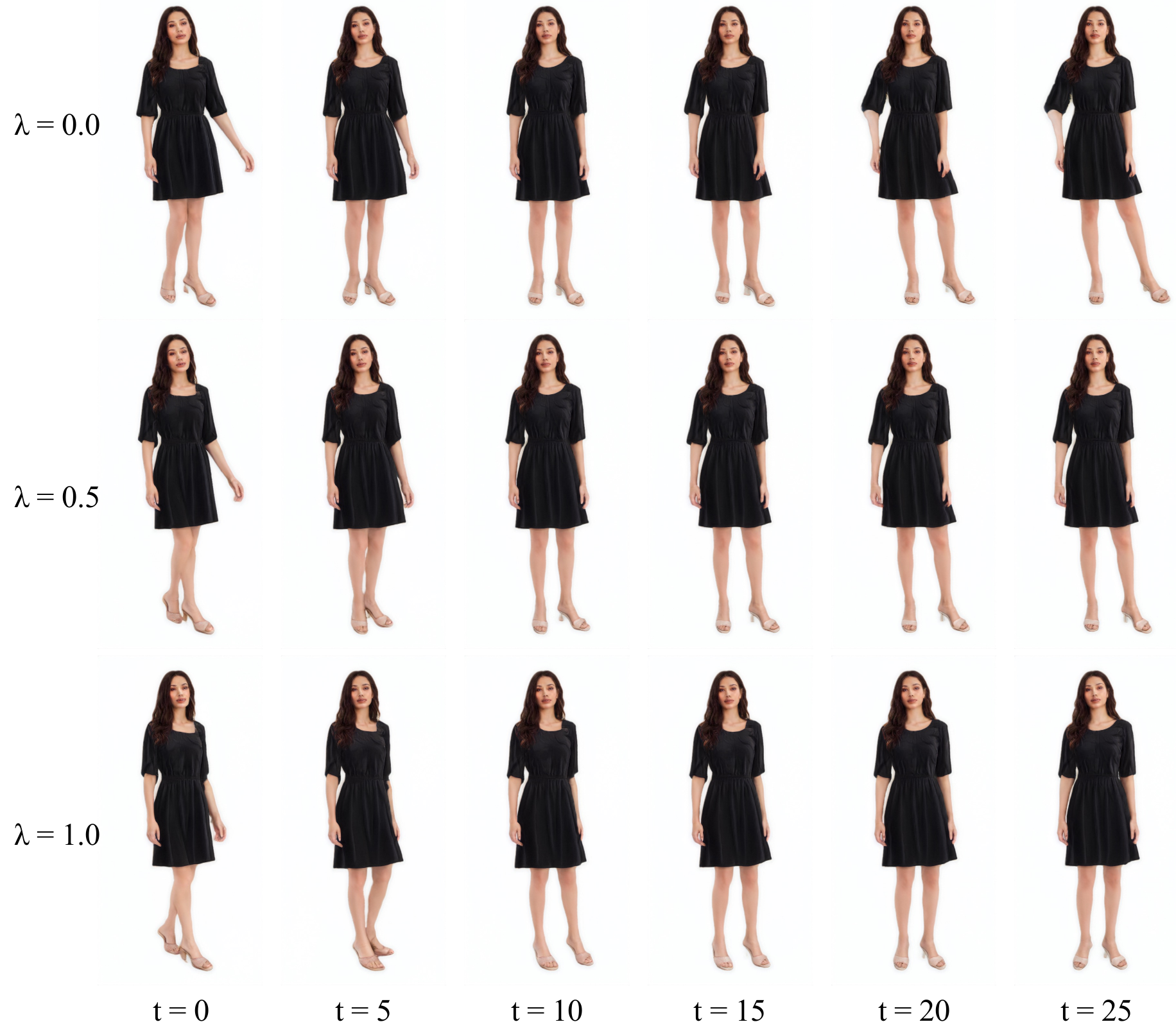}
    \caption{Obtaining the dynamic representation from two videos, we interpolate between them with values $\lambda=0.0$, $\lambda=0.5$, and $\lambda=1.0$, and show that the dynamics do change smoothly as we interpolate. We interpolate over 25 frames.}
    \label{fig:dynamic_interpolation}
\end{figure*}

\begin{figure*}[!t]
    \centering
    \includegraphics[width=\linewidth]{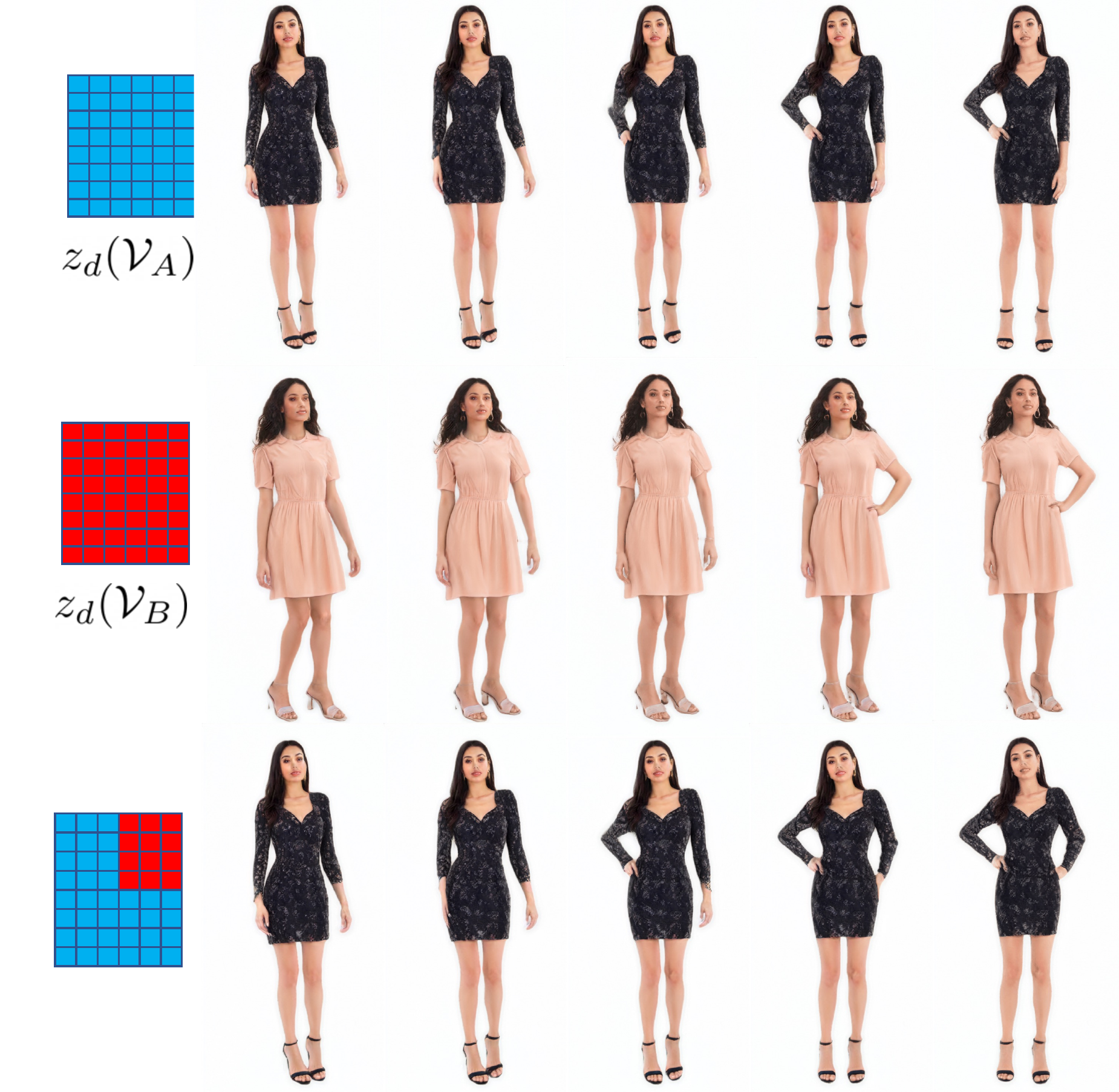}
    \caption{We transfer the upper right dynamics from $\mathcal{V}_{B}$ to $\mathcal{V}_{A}$, while keeping the rest of the dynamics from $\mathcal{V}_{A}$. This results in the right arm moving upwards, while the rest of the dynamics are unchanged.}
    \label{fig:local_dyn_editing}
\end{figure*}

\begin{figure*}[!t]
    \centering
    \includegraphics[width=0.8\linewidth]{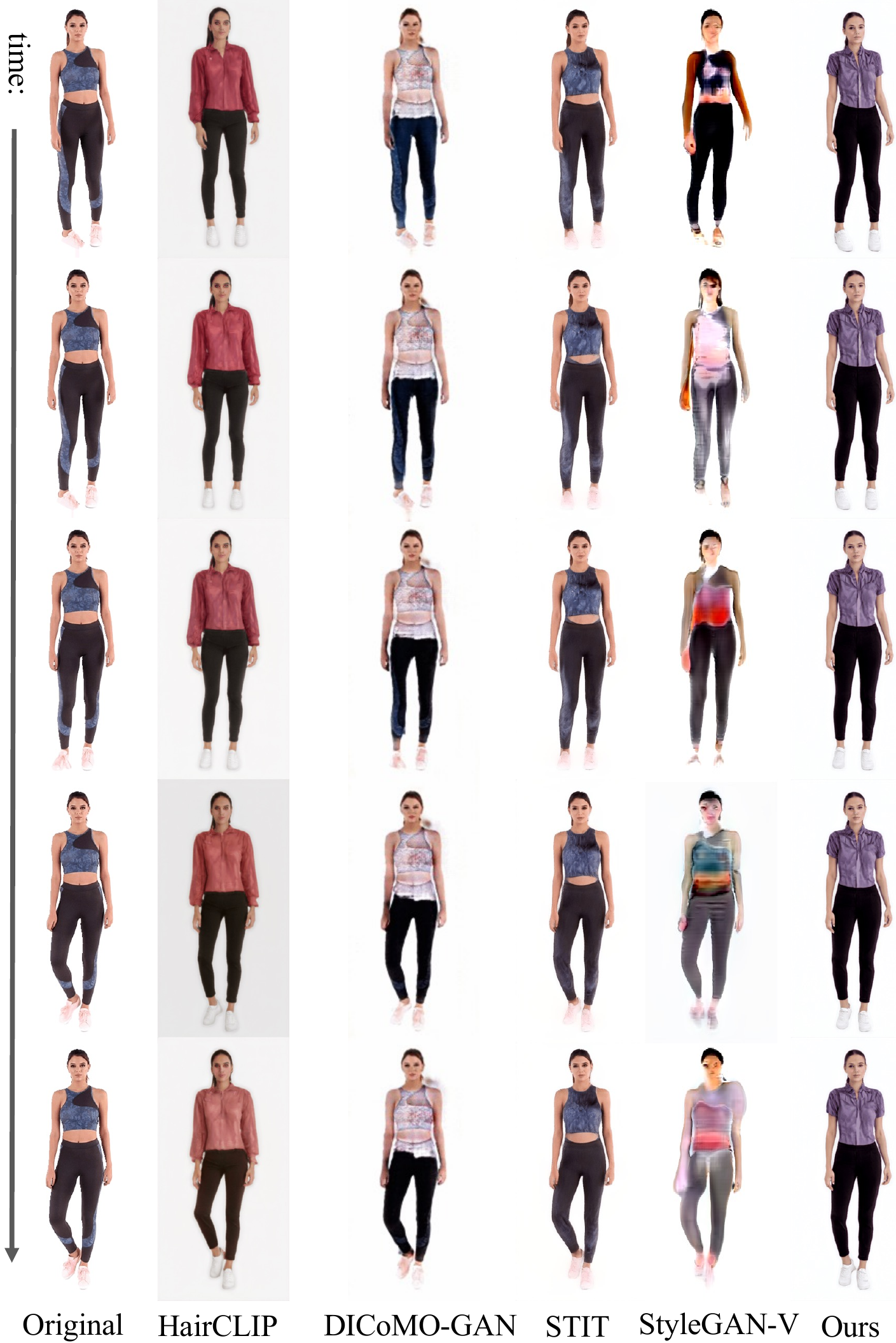}
    \caption{Here, we perform additional manipulations using the baselines. We exclude the latent transformer results since it is unable to perform complex manipulations without multiple steps. The source text is ``\textit{a photo of a woman wearing a crop top}", and the target text is ``\textit{a photo of a woman wearing a \textbf{blouse}}".}
    \label{fig:baselines_supp}
\end{figure*}

\begin{figure*}[!t]
    \centering
    \includegraphics[width=\linewidth]{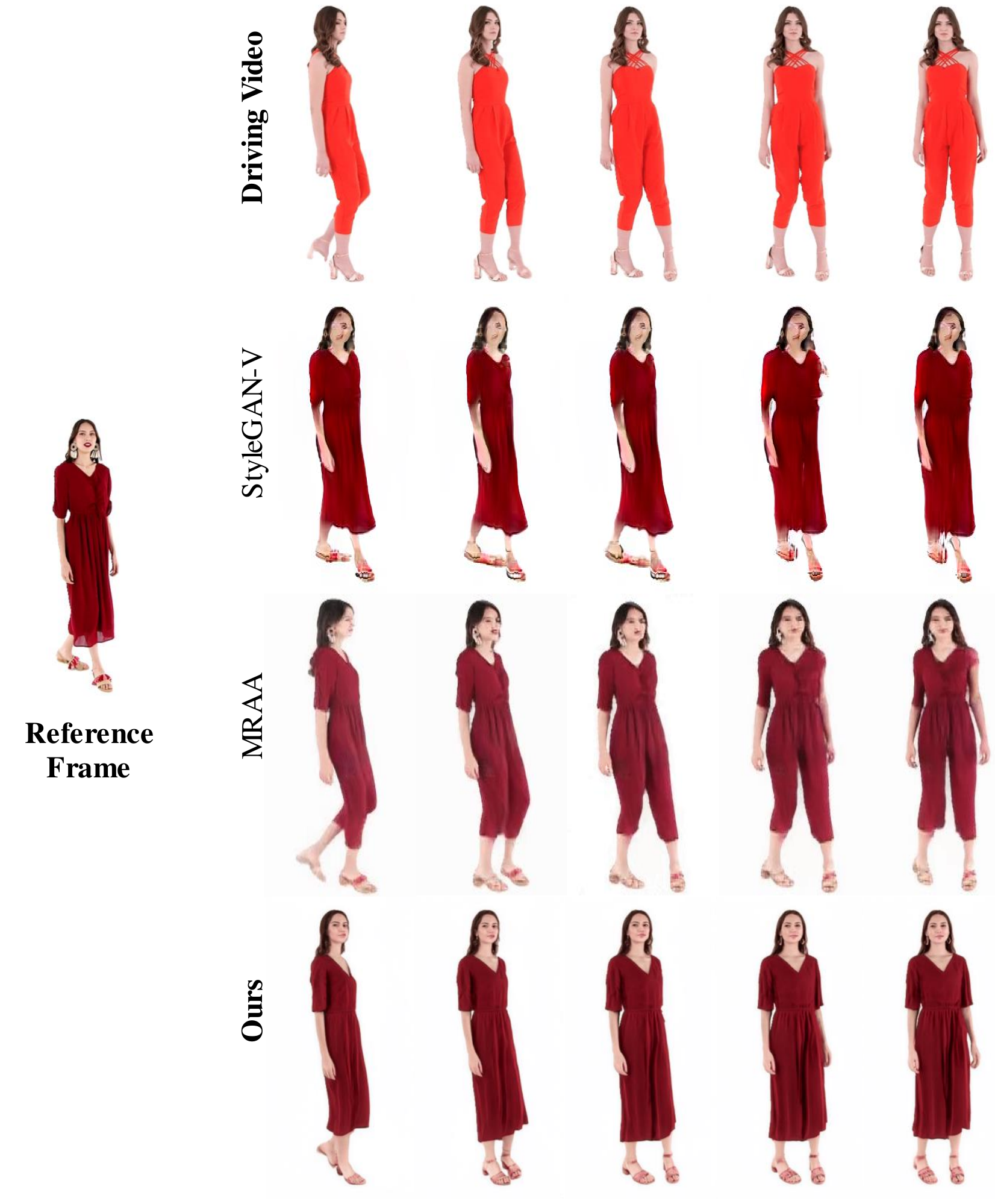}
    \caption{We perform image animation using our model and multiple baselines. We obtain the dynamics from the driving video in the first row and apply it to the reference frame to generate the videos. Our results obtain the highest perceptual quality, while also matching the dynamics of the driving video, and structure of the reference frame. MRAA modifies the structure according to the driving video, instead of just transferring motion, while StyleGAN-V has some motion transferred, but has a very low perceptual quality.}
    \label{fig:driving1}
\end{figure*}

\begin{figure*}[!t]
    \centering
    \includegraphics[width=\linewidth]{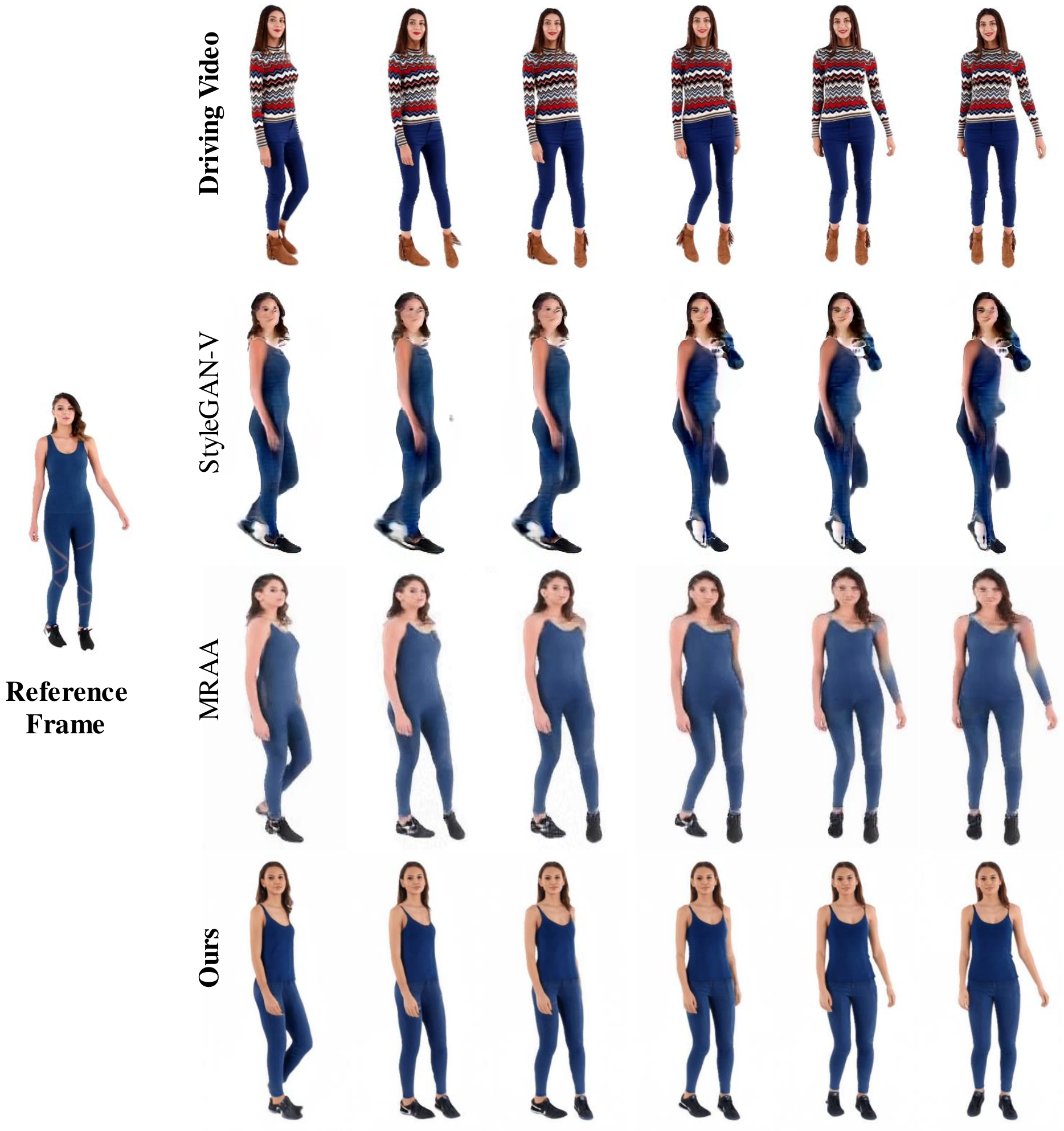}
    \caption{Another example on image animation. Our model produces the result with the highest perceptual quality while being consistent with the driving video and reference frame. MRAA has some inconsistencies in the arm, and StyleGAN-V is unable to capture the motion in any meaningful way.}
    \label{fig:driving2}
\end{figure*}

\begin{figure*}[!t]
    \centering
    \includegraphics[width=0.625\linewidth]{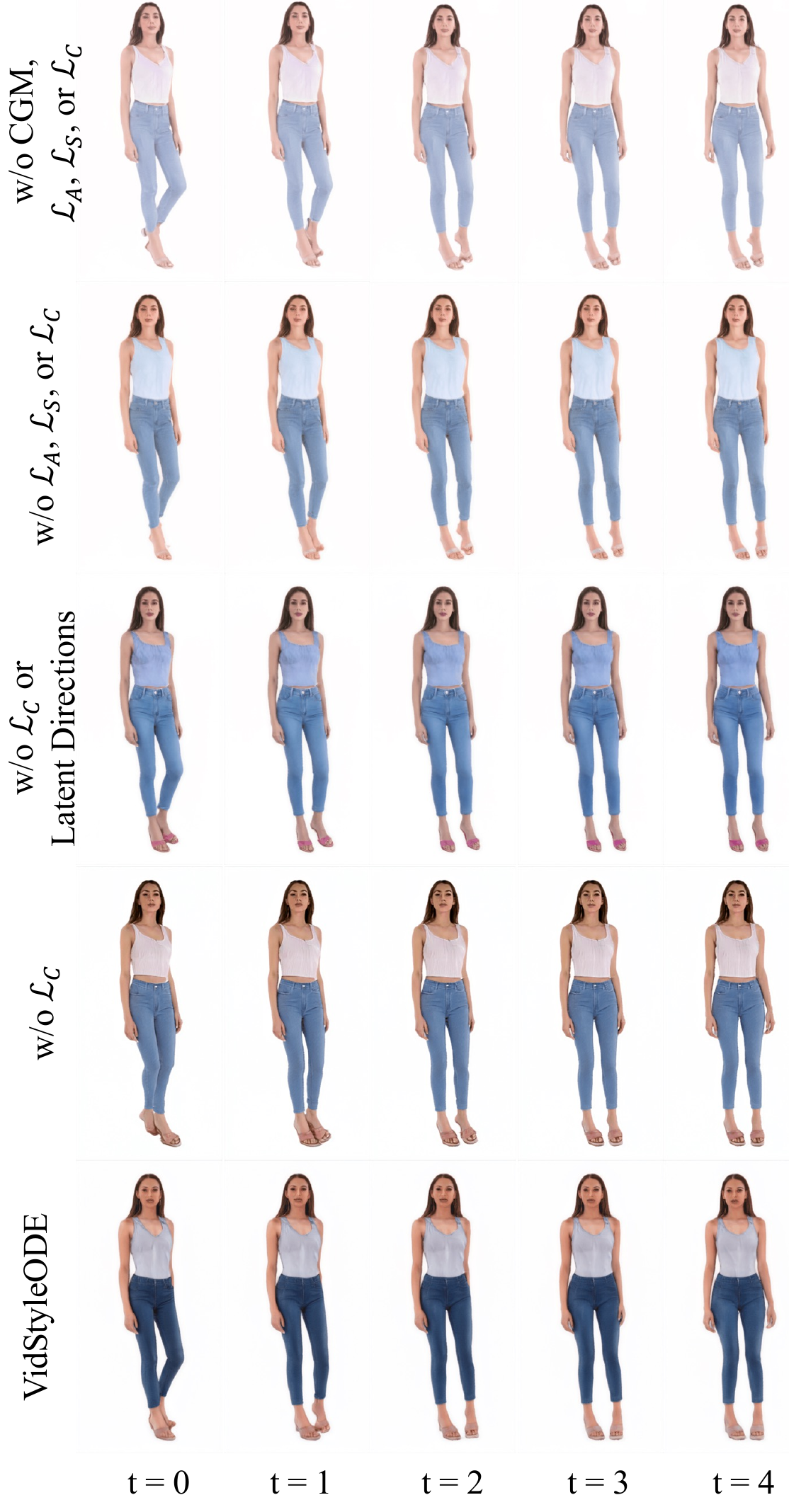}
    \caption{We provide examples to support our ablation study. The first row is the model without the conditional generative model $f_G$, structural loss, appearance loss, or consistency loss. The second row is without structural loss, appearance loss, or consistency loss. The third row is without consistency loss, and without using directions. The fourth row is just without consistency loss. Finally, the fifth row is our best model, with everything.}
    \label{fig:ablation}
\end{figure*}

\subsection{RAVDESS dataset}
\begin{itemize}[noitemsep,leftmargin=*,topsep=0pt]
\item In~\cref{fig:video_manipulation_3}, we present sample text-guided editing examples on a sample video from the RAVDESS dataset for three different target texts. Our proposed \name method accurately manipulates the provided source videos in a temporally-consistent way according to the provided target text. It successfully changes the eye color, the hair color, and the gender of the person of interest.
\item \cref{fig:interpolation2} provides example frame interpolation results over the provided face frames. As seen, our \name method accurately predicts what the frames from the missing timestamps look like. 
\item In~\cref{fig:extrapolation_1_frame2}, we show that our \name method can also animate still frames via extrapolation.
\item Finally, in~\cref{fig:baselines_2}, we give qualitative comparisons against state-of-the-art editing techniques on a sample video having the source description ``\textit{A woman with blond hair, and green eyes}” and with the target description being specified as ``\textit{A woman with brown hair and blue eyes}”. As seen, compared to the state-of-the-art methods, \name generates a temporally coherent output depicting all the proper edits done on the source video.
\end{itemize}

\begin{figure*}[!t]
    \centering
    \includegraphics[width=\linewidth]{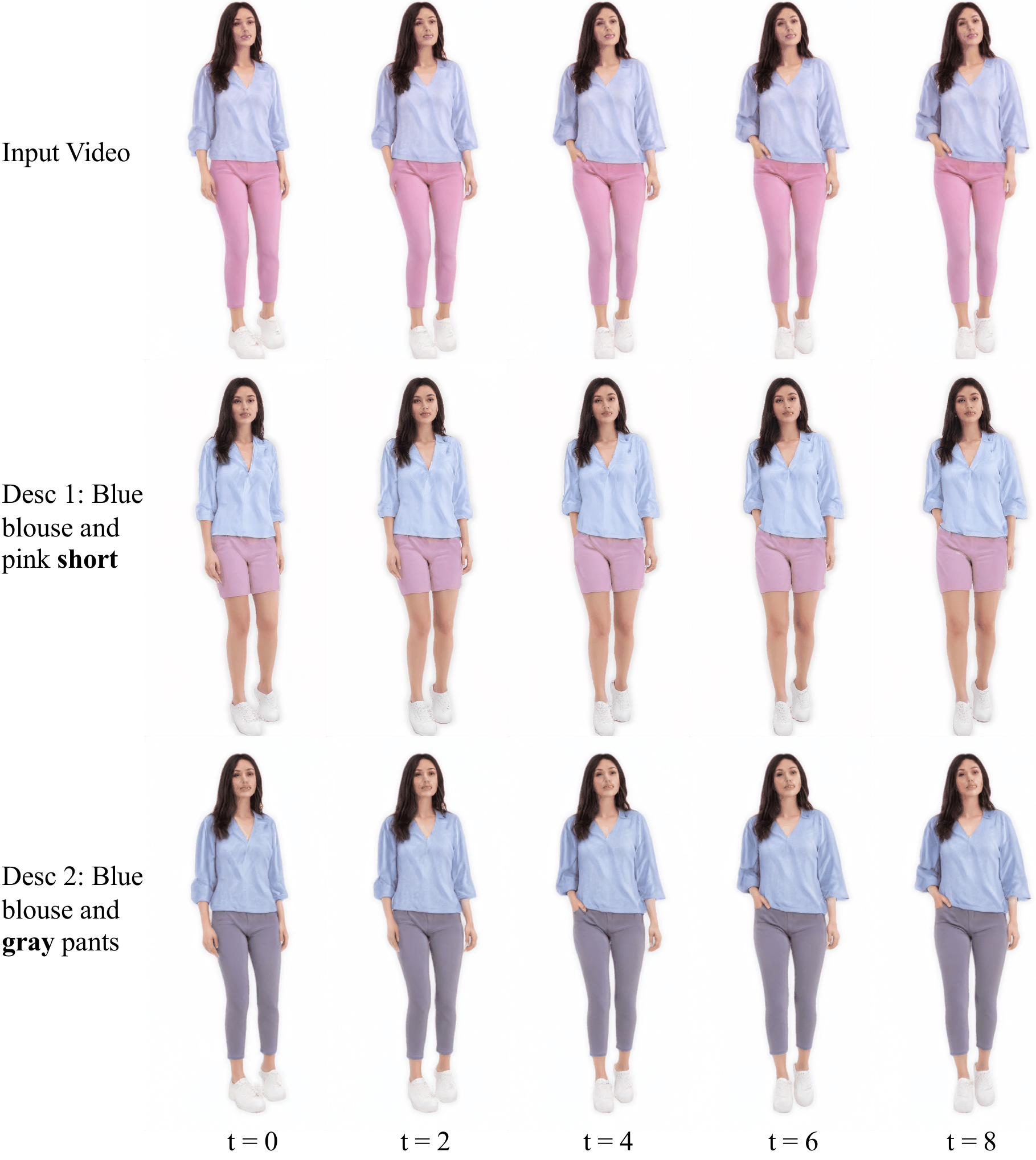}
    \caption{We perform two different manipulations to a sample video (the source video) from the Fashion Videos dataset and display the corresponding results here. Target 1 uses the target text ``\textit{A photo of a woman wearing blue blouse and pink \textbf{short}}". Target 2 uses the target text ``\textit{A photo of a woman wearing blue blouse and \textbf{gray} pants}".}
    \label{fig:video_manipulation}
\end{figure*}
\begin{figure*}[!t]
    \centering
    \includegraphics[width=\linewidth]{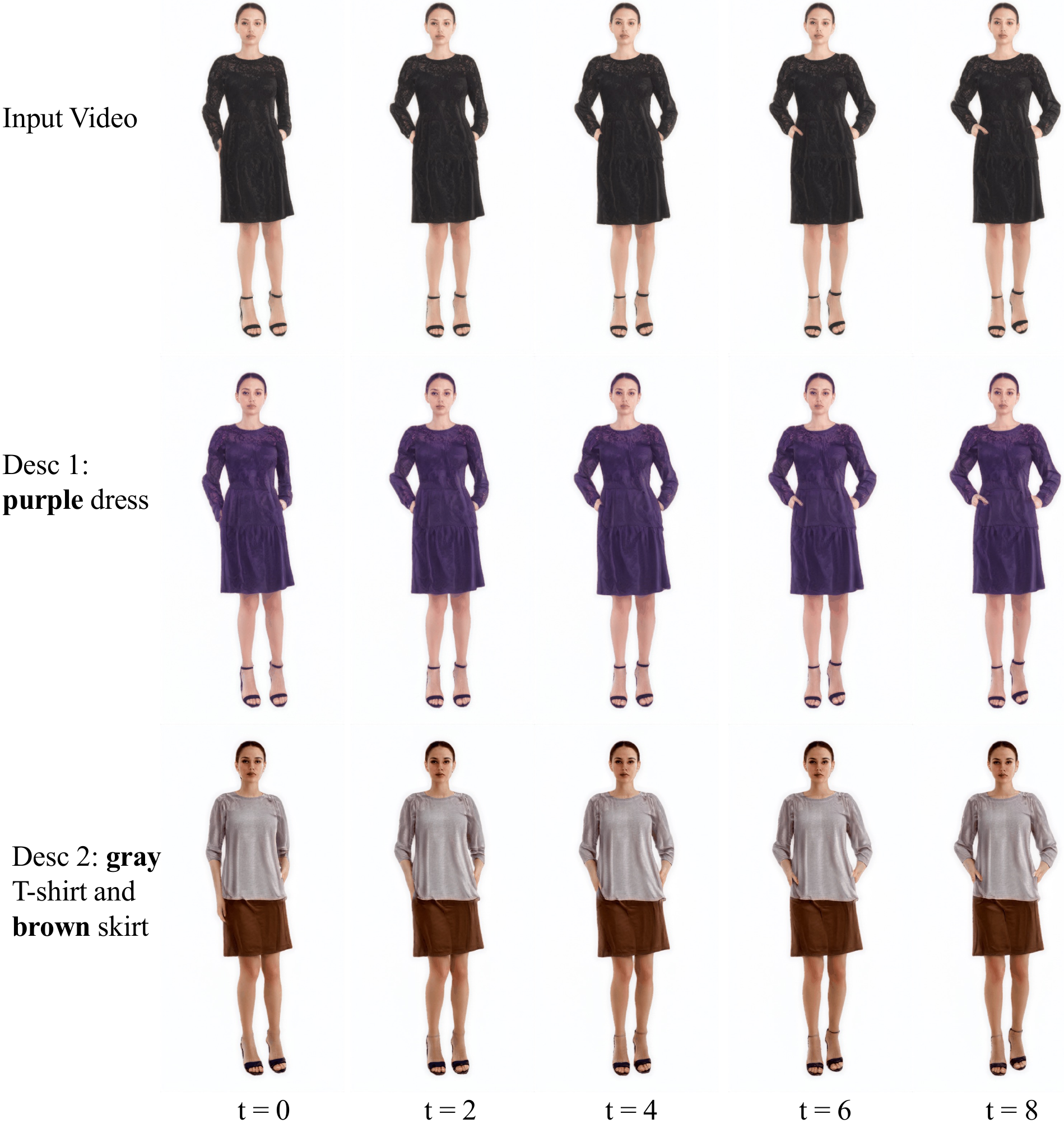}
    \caption{We perform two different manipulations to a sample video (the source video) from the Fashion Videos dataset and display the corresponding results here. Target 1 uses the target text ``\textit{A photo of a woman wearing a \textbf{purple} dress}". Target 2 uses the target text ``\textit{A photo of a woman wearing \textbf{gray} T-shirt and \textbf{brown} skirt.}".}
    \label{fig:video_manipulation_2}
\end{figure*}

\begin{figure*}[!t]
    \centering
    \includegraphics[width=\linewidth]{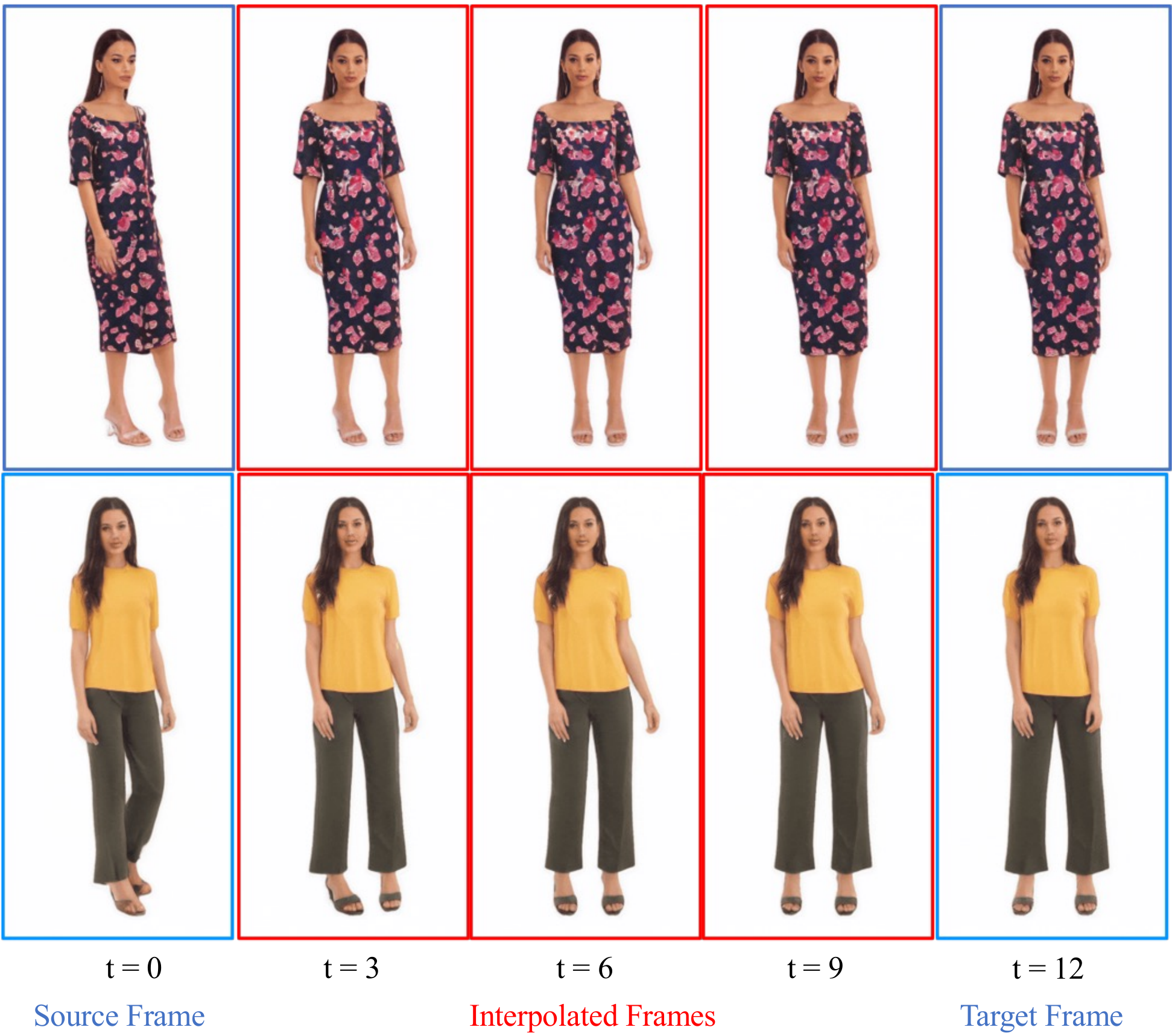}
    \caption{To perform interpolation, we provide the first and last frames (shown in blue) to the model and then generate the whole video. We display 3 evenly-spaced interpolated frames for each video (shown in red).}
    \label{fig:interpolation}
\end{figure*}

\begin{figure*}[!t]
    \centering
    \includegraphics[width=\linewidth]{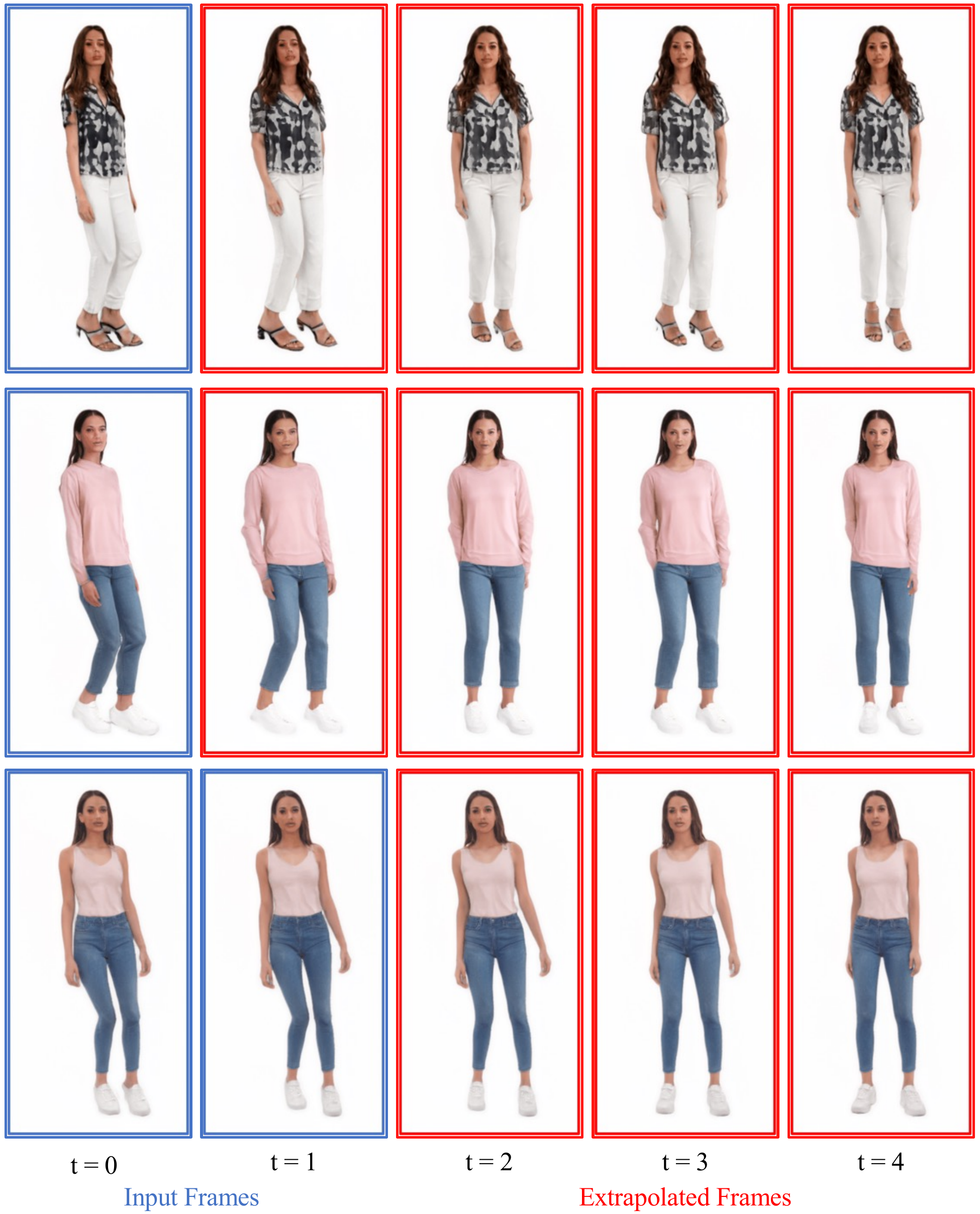}
    \caption{To perform extrapolation from a single frame, we provide just the initial frame (shown in blue), and then generate the next 4 frames (shown in red).}
    \label{fig:extrapolation_1_frame}
\end{figure*}

\begin{figure*}
    \centering
    \includegraphics[width=\linewidth]{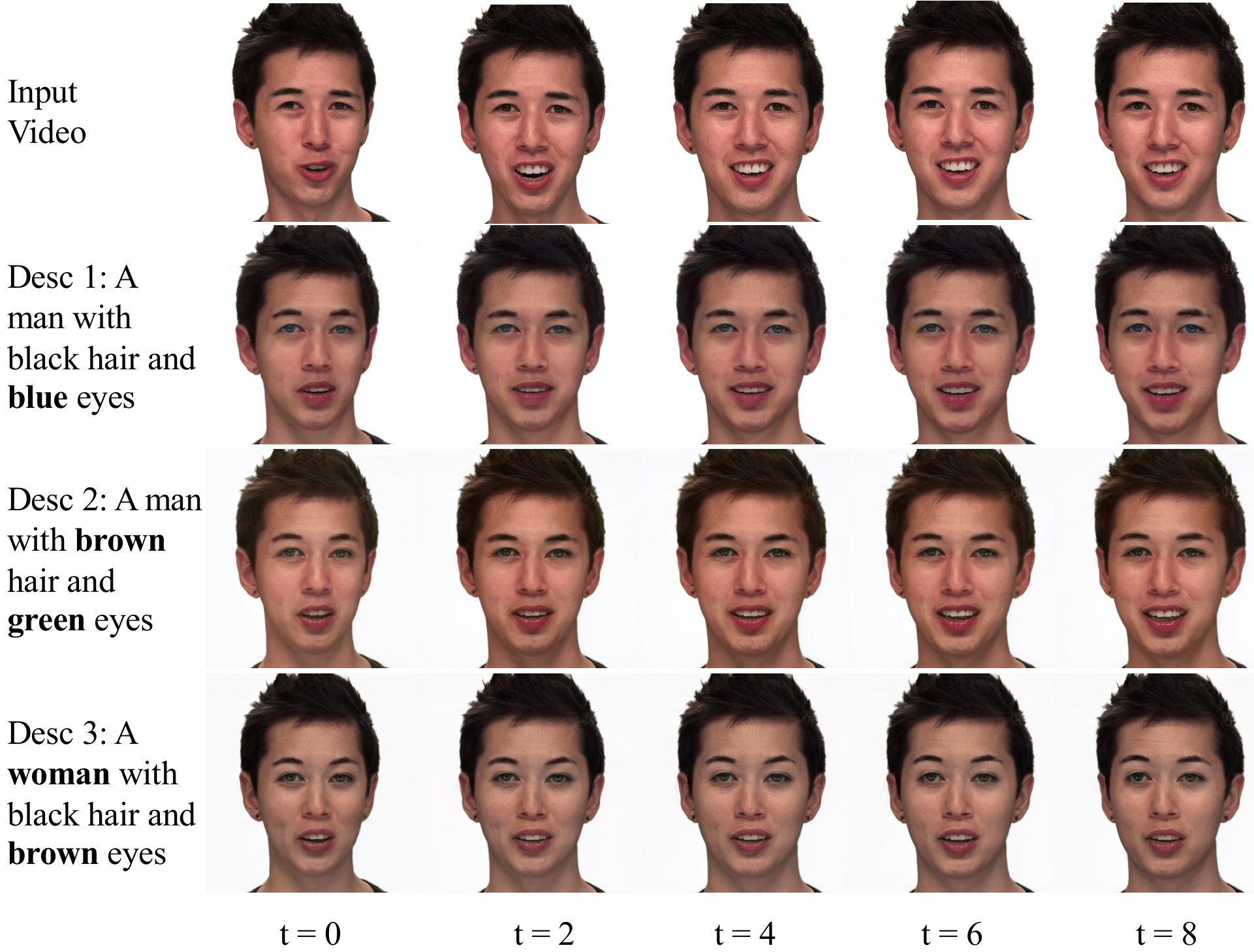}
    \caption{We perform three different manipulations to a sample video (the source video) from the RAVDESS  dataset, and display the corresponding results here. Target 1 uses the target text ``\textit{A man with black hair and \textbf{blue} eyes}". Target 2 employs the target text ``\textit{A man with \textbf{brown} hair and \textbf{green} eyes.}". Target 3 uses the target text ``\textit{A \textbf{woman} with black hair and brown eyes."}.}
    \label{fig:video_manipulation_3}
\end{figure*}

\begin{figure*}[!t]
    \centering
    \includegraphics[width=\linewidth]{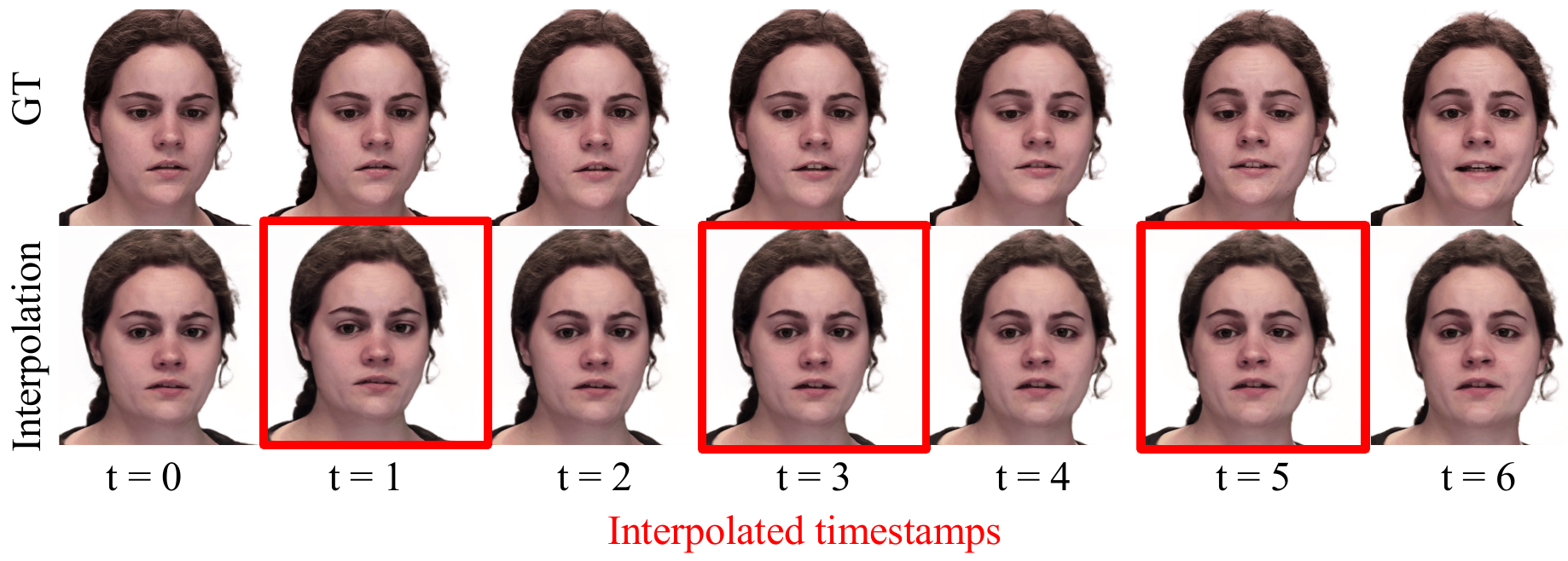}
    \caption{To perform interpolation, we provide four distinct frames with different timestamps ($t=0$, $t=2$, $t=4$, and  $t=6$) (shown in blue) to the model, and then generate the unobserved frames for timestamps $t=1$, $t=3$, and $t=5$ (shown in red).}
    \label{fig:interpolation2}
\end{figure*}
\begin{figure*}[!t]
    \centering
    \includegraphics[width=\linewidth]{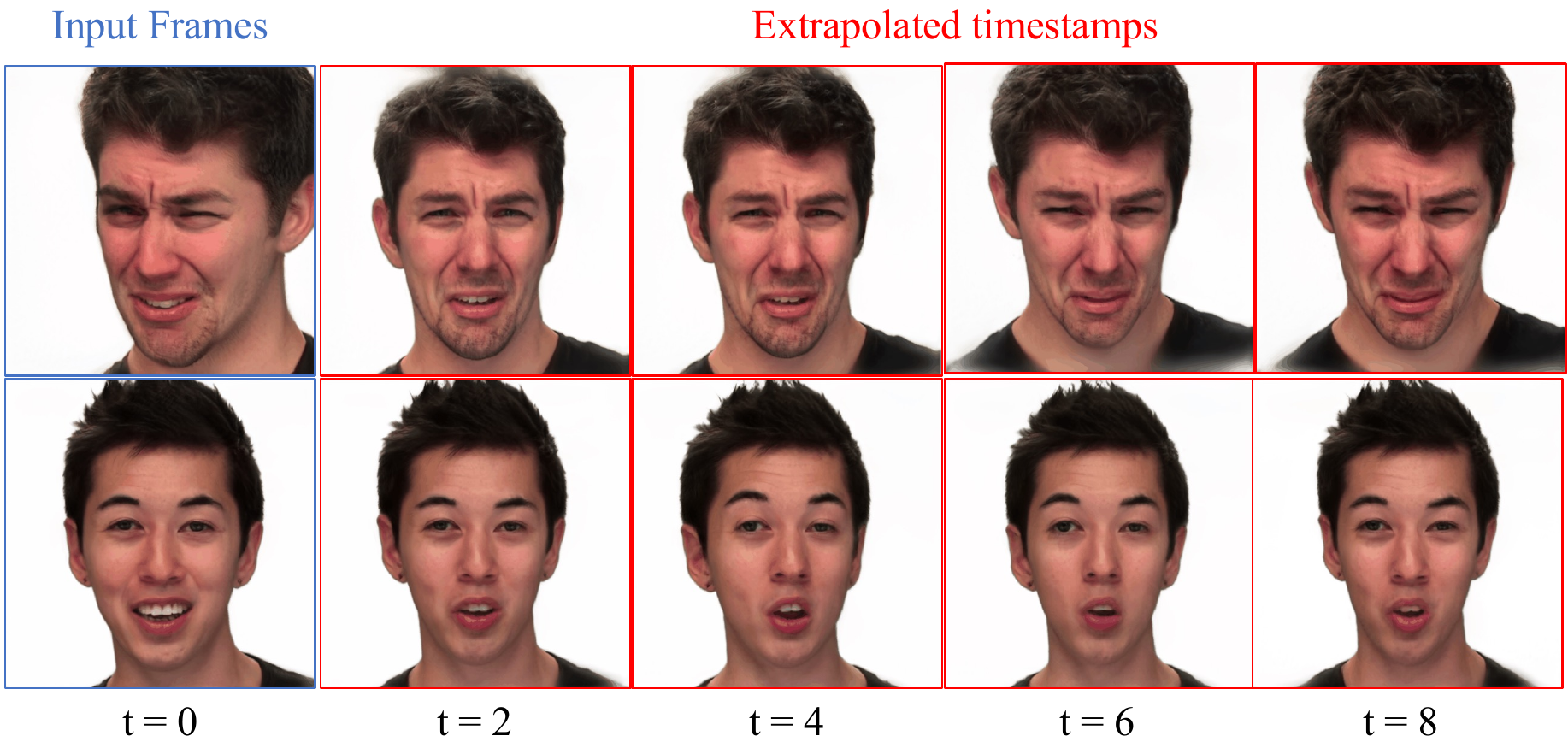}
    \caption{Extrapolation from a single frame: we provide just the initial frame (shown in blue), and then generate the next 4 frames (shown in red).}
    \label{fig:extrapolation_1_frame2}
\end{figure*}

\begin{figure*}[!t]
    \centering
    \includegraphics[width=\linewidth]{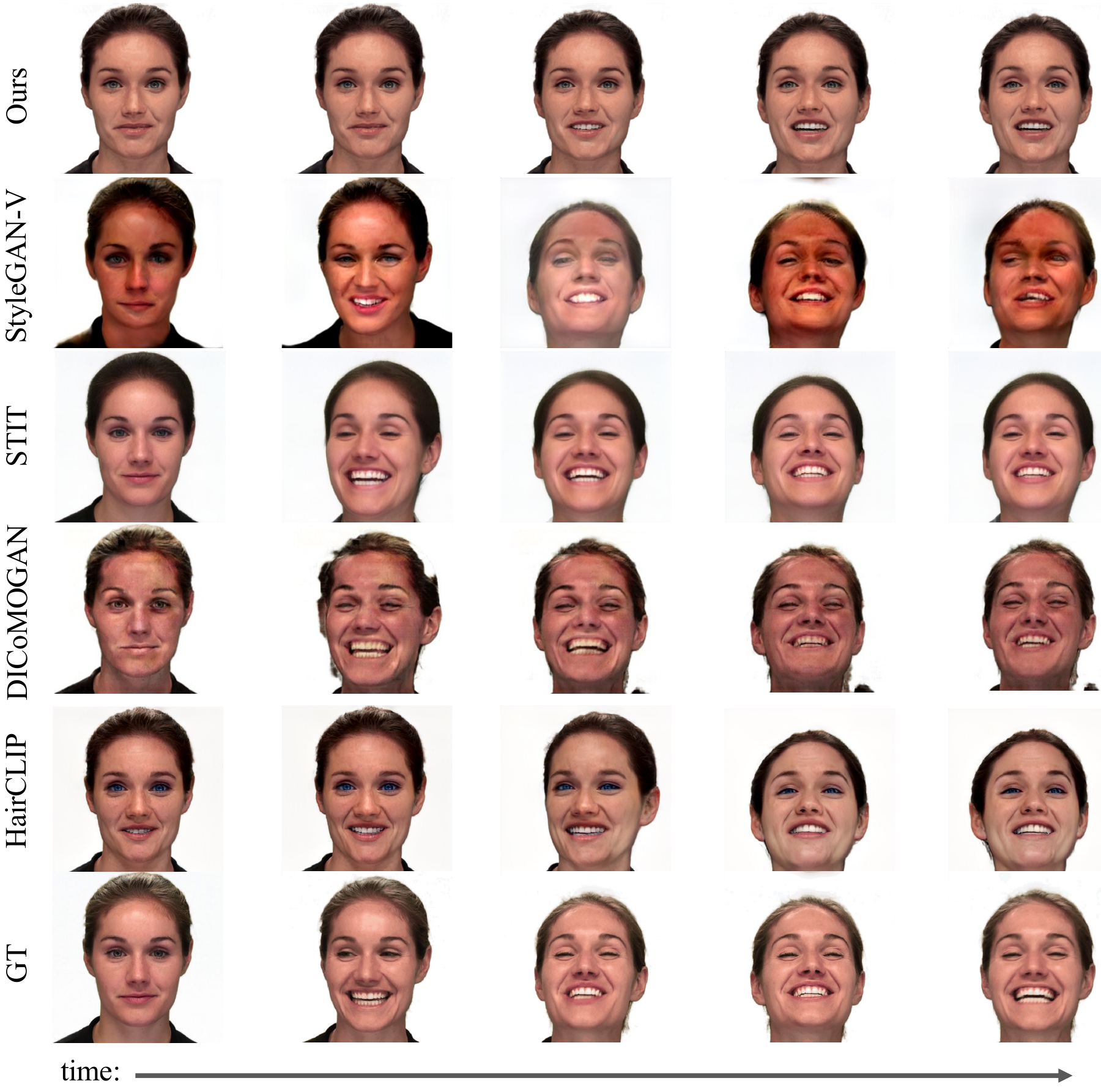}
    \caption{Qualitative results of our approach and the competing editing methods. The description of the source image is ``\textit{A woman with blond hair, and green eyes}”, while the target description is specified as ``\textit{A woman with \textbf{brown} hair and \textbf{blue} eyes}”.}
    \label{fig:baselines_2}
\end{figure*}

 \fi

\end{document}